\documentclass{article}
\usepackage[preprint]{neurips_2025}

\usepackage{natbib}
\usepackage{graphicx}
\usepackage[utf8]{inputenc} 
\usepackage[T1]{fontenc}    
\usepackage{hyperref}       
\usepackage{url}            
\usepackage{booktabs}       
\usepackage{amsfonts}       
\usepackage{amsmath}
\usepackage{amsthm}
\usepackage{amssymb}
\usepackage{nicefrac}       
\usepackage{microtype}      
\usepackage{xcolor}         
\usepackage[font=small,labelfont=bf]{caption}
\usepackage{enumitem}
\usepackage{wrapfig}
\usepackage{listings}
\usepackage{caption}
\usepackage{multirow}
\usepackage[most]{tcolorbox}
\usepackage[normalem]{ulem}
\usepackage{xspace}
\usepackage{float}
\usepackage{tabularx}
\usepackage{booktabs}
\usepackage[normalem]{ulem}
\usepackage{xcolor} 
\usepackage{soul}   
\usepackage{svg}
\usepackage{array}
\usepackage{longtable}
\usepackage{adjustbox}
\usepackage{makecell}
\usepackage{pifont}
\newcommand{\cmark}{\ding{51}}%
\newcommand{\xmark}{\ding{55}}%

\useunder{\uline}{\ul}{}

\newtcolorbox{systempromptbox}[1][]{%
    colback=orange!5!white, 
    colframe=orange!30!orange, 
    fonttitle=\bfseries,
    title=Magistral's system prompt,
    sharp corners,
    #1
}

\definecolor{mygreybg}{gray}{0.95}

\sethlcolor{mygreybg}

\newcommand{\greyblockmono}[1]{%
  {\texttt{\hl{#1}}}%
}

\title{Magistral}

\begin{document}

\maketitle

\vspace{-0.1in}
\begin{center}
\vspace{-45pt}
\centering
\includegraphics[width=0.8\linewidth,keepaspectratio]{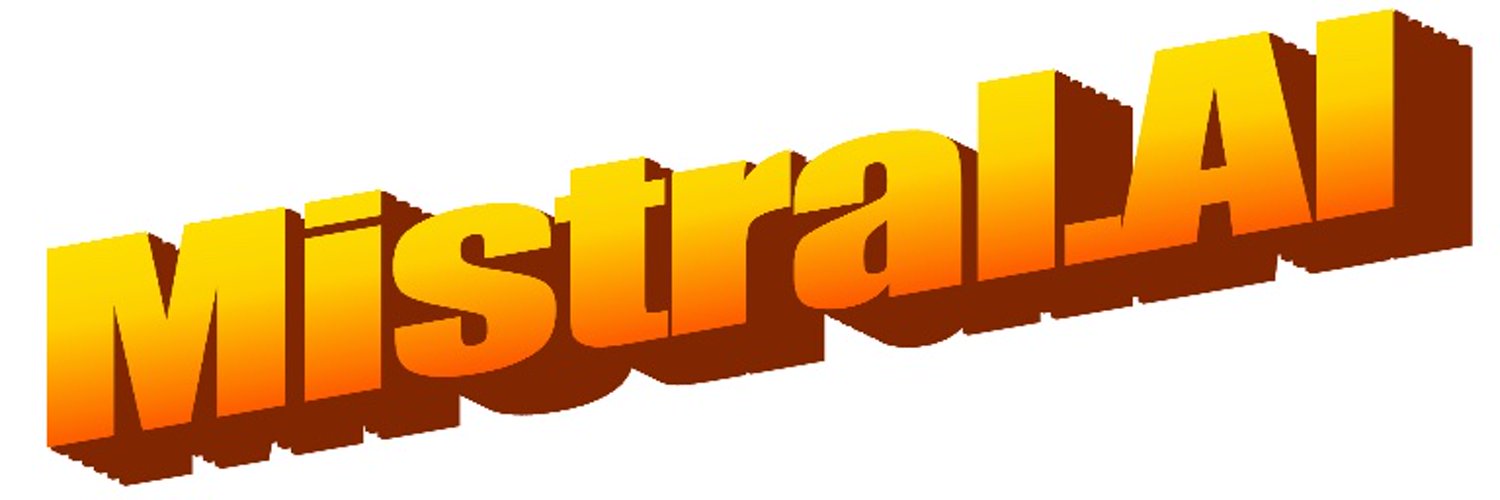}
\end{center}

\begin{abstract}
We introduce Magistral, Mistral’s first reasoning model and our own scalable reinforcement learning (RL) pipeline. Instead of relying on existing implementations and RL traces distilled from prior models, we follow a ground up approach, relying solely on our own models and infrastructure. 
Notably, we demonstrate a stack that enabled us to explore the limits of pure RL training of LLMs, present a simple method to force the reasoning language of the model, and show that RL on text data alone maintains most of the initial checkpoint's capabilities. We find that RL on text maintains or improves multimodal understanding, instruction following and function calling.
We present  Magistral Medium, trained for reasoning on top of Mistral~Medium~3 with RL alone, and we open-source Magistral Small (Apache 2.0)  which further includes cold-start data from Magistral Medium.

\end{abstract}
\vspace{-0.3cm}

\section{Introduction}

Enhancing the reasoning abilities of large language models (LLMs) has emerged as a key frontier in modern AI research. 
Reasoning models such as o1~\citep{o12024openai} differ widely from classic chatbots, leveraging longer chains-of-thought to improve performance on complex tasks. 
The seminal work by \cite{deepseekai2025deepseekr1} gave the community crucial insights on the Reinforcement Learning from Verifiable Rewards~(RLVR) recipe, for creating reasoning models at scale.

\looseness=-1 In this paper, we introduce Mistral's first reasoning models: Magistral Small and Magistral Medium, based on the Mistral~Small~3 and Mistral~Medium~3 models respectively, and outline our proposed RLVR framework in detail.
The key contributions of our paper are the following:
\begin{itemize}
    \item We present in detail how we trained Magistral Medium with RL alone, with no distillation from pre-existing reasoning models, yielding a nearly 50\% boost in AIME-24 (pass@1). 
    \item We discuss in depth the infrastructure and design choices that enable large-scale online RL. Our asynchronous system enables fast, continuous RL training by updating generators frequently without interrupting them, balancing efficiency with on-policyness.
    \item We present a simple yet effective strategy to make the model multilingual, where both the chain-of-thought and the final response are written in the user's language.
    \item \looseness=-1 We contribute insights that add to, or contradict, existing RLVR literature, for example on whether RL can improve upon the distillation SFT baseline for small models. We also show that multimodal reasoning capabilities emerge with online RL with textual data on top of a multimodal model. We share the results of our unsuccessful experiments.
    \item We release the weights of Magistral Small (24B) under the Apache 2 license\footnote{https://huggingface.co/mistralai/Magistral-Small-2506}.
\end{itemize}

\begin{figure}[t]
    \centering
    \makebox[\textwidth][c]{
        \includegraphics[width=\textwidth]{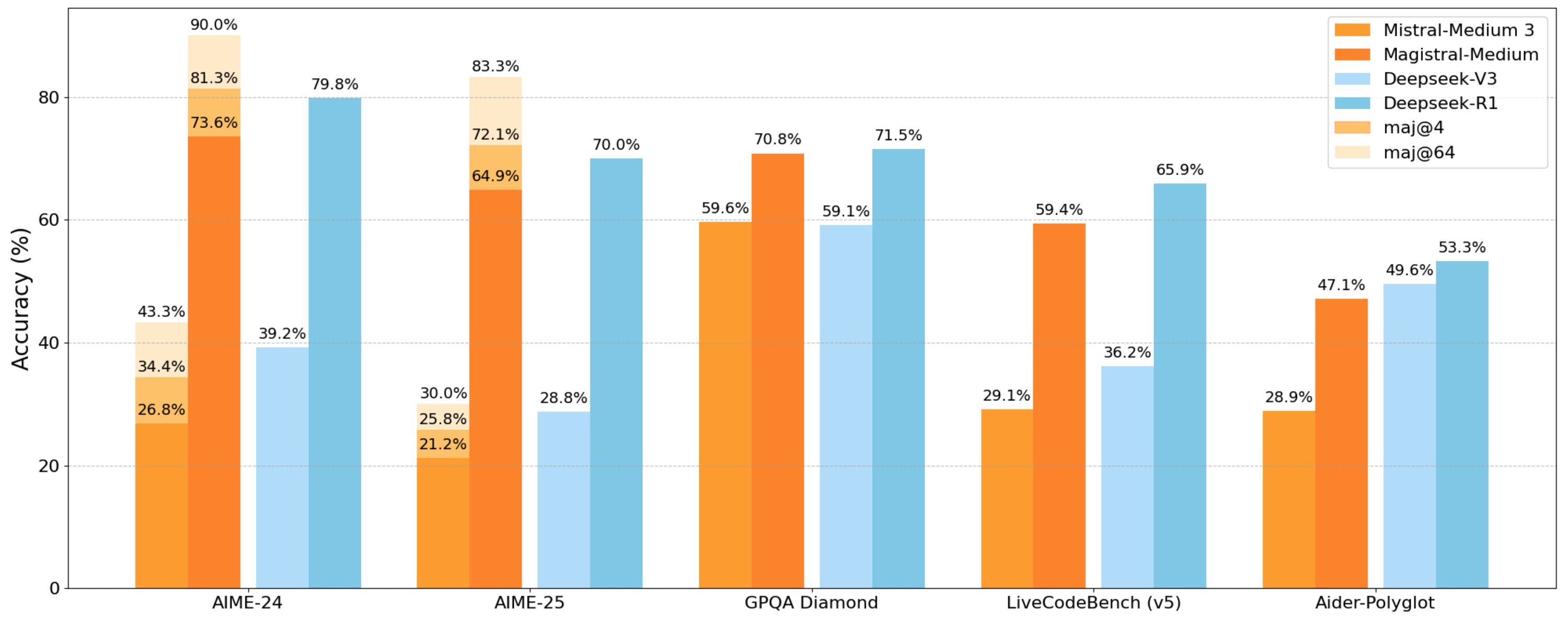}
    }
    \caption{
    \looseness=-1 \textbf{Performance of Magistral Medium on common reasoning benchmarks.}
    We highlight the strength of our proposed RLVR framework, which yields a 50\% increase in AIME-24 (pass@1) over the initial Mistral Medium 3 checkpoint, \textit{without any cold-start reasoning traces}. 
    We compare against analogous results from~\citep{deepseekai2025deepseekr1}, which show RL improvements from DeepSeek-v3 to DeepSeek-R1 (January 25).
    Magistral Medium reaches 90\% accuracy on AIME-24 with majority voting.
    \label{fig:magistral-r1}
    }
    \vspace{-0.2cm}
\end{figure}

\looseness=-1 The paper is organized as follows: Section~\ref{sec:method} details the RL algorithm we used, along with the design choices implemented to guide the reasoning models in terms of language and format; Section~\ref{sec:infra} presents our scalable infrastructure that supports efficient training on a large cluster of GPUs; Section~\ref{sec: data} discusses the data selection process we employed for efficient and effective training; Section~\ref{sec:results} presents the performance of Magistral on reasoning and multilingual benchmarks; Section~\ref{sec:ablations} shows the ablations done to motivate the training choices; Section~\ref{sec:analysis} presents a PCA-based study of the model weights' trajectory during RL, demonstrates that RL on text data preserves or even improves multimodal capabilities, and includes methods that worked poorly for Magistral; Section~\ref{sec:rl_oss} shows that one can train a model to perform on par with R1 with distillation followed by RL, which we did not use for Magistral Medium; Finally, we conclude with some future directions in Section~\ref{sec:conclusion}.

\vspace{-0.2cm}
\section{Methodology}
\label{sec:method}

\vspace{-0.2cm}

In this section, we outline the training methodology used to develop the Magistral models. This includes our optimizations of the GRPO algorithm for training stability (Section~\ref{subsec: rl algo}) and our training reward to improve both mathematical and coding capabilities, while ensuring the model adheres to proper format, length, and language usage (Section~\ref{subsec:reward}).

\subsection{Reinforcement learning algorithm}
\label{subsec: rl algo}

\looseness=-1 We use Group Relative Policy Optimization (GRPO)~\citep{shao2024deepseekmathpushinglimitsmathematical} as our RL algorithm. Unlike PPO~\citep{schulman2017proximal}, GRPO eliminates the need for a `critic model', and instead uses the average reward from multiple generations per prompt from the policy to compute a baseline for advantage calculation. Specifically, GRPO optimizes the policy $\pi_{\theta}$ to maximize the following objective:

\begin{align*}
\mathcal{J}_{\text{GRPO}}(\theta) &= \mathbb{E}_{q \sim P(Q), \{o_i\}_{i=1}^G \sim \pi_{\theta_{\text{old}}}(\cdot|q)} \notag \\  \Bigg[ & \sum_{i=1}^{G} \sum_{t=1}^{|o_i|} \frac{1}{|o_i|} \Big( \min\left[\frac{\pi_\theta(o_{i,t}|q, o_{i,<t})}{\pi_{\theta_{\text{old}}}(o_{i,t}|q, o_{i,<t})} \hat{A}_{i,t}, \; \text{clip}(\frac{\pi_\theta(o_{i,t}|q, o_{i,<t})}{\pi_{\theta_{\text{old}}}(o_{i,t}|q, o_{i,<t})}, 1-\varepsilon, 1+\varepsilon)\hat{A}_{i,t}\right] \notag \\
 -  & \beta D_{\mathrm{KL}}[\pi_\theta(\cdot|q) \| \pi_{\text{ref}}(\cdot|q)] \Big) \Bigg], 
\end{align*}
where $q$ represents queries drawn from the input dataset, $o$ represents the generation of the model, \(\varepsilon\) is the PPO clipping threshold, \(\beta\) is the KL penalty coefficient, and \(D_{\mathrm{KL}}\) denotes the Kullback–Leibler divergence between the current policy \(\pi_\theta\) and the reference policy \(\pi_{\text{ref}}\).
\looseness=-1 The relative advantage, or the group normalized advantage, is given by $\hat{A}_{i,t} = \frac{r_i - \mu}{\sigma}$ where \(\mu\) and \(\sigma\) are the mean and standard deviation of rewards computed within a single group. Building on prior work adapting GRPO for reasoning tasks ~\citep{yu2025dapo, liu2025understanding, hu2025openreasonerzeroopensourceapproach}, we introduced several modifications:

\textbf{Eliminating KL divergence.} The KL divergence penalty constrains the online policy from deviating too much from a reference policy, helping to maintain alignment with the initial model. However, in GRPO, the policy diverges substantially regardless, and maintaining a copy of the reference model for KL computation incurs a compute cost we find unjustified. We remove the KL penalty entirely.

\textbf{Loss normalization.} To avoid introducing length biases between generations in one group, we normalize the loss by first adding token-wise loss for all tokens and all generations and then dividing by the total length of generations in the group $\sum_{i=1}^G |o_i|$.

\textbf{Advantage normalization.} We estimate the advantage of each token simply as 
$\hat{A}_{i, t} = \hat{A}_i = r_i - \mu$, where $\mu$ is the mean of rewards within a group.
Following ~\cite{andrychowicz2020}, we additionally normalize the advantages in each minibatch as \smash{$\hat{A}_{i, t}^{\text{norm}} = (\hat{A}_i - \hat{A}^{\text{mean}}) / \hat{A}^{\text{std}}$} where $\hat{A}^{\text{mean}}$ and $\hat{A}^{\text{std}}$ are the sequence-wise mean and standard deviation of the advantages $\hat{A}_i$ in a minibatch.

\textbf{Relaxing the trust region's upper bound.} We allow the model to explore rare but potentially insightful reasoning steps, preventing deterministic policies. We adopt the \textit{Clip-Higher} \citep{yu2025dapo} strategy to address entropy collapse. In standard GRPO, $\varepsilon$-clipping limits exploration by restricting the increase in probability of low-likelihood tokens, hindering the reinforcement of rare but important reasoning paths. By increasing the upper clipping threshold to $\varepsilon_{\text{high}}$, low-probability tokens have more room to grow, enhancing entropy and diversity in outputs, and improving reasoning exploration. We found that careful tuning of $\varepsilon_{\text{high}}$ is crucial to maintaining stability in the RL run. We adjusted it between 0.26 and 0.28 during the training to keep the group entropy stable.

\textbf{Eliminating non-diverse groups.} Groups where all generations are either entirely correct or wrong have zero advantage and therefore contribute nothing to the batch loss. This results in smaller gradients with increased noise sensitivity. To address this, we filter out all groups with zero advantage when forming training batches.

The final GRPO loss with all modifications highlighted in red is
\begin{align*}
\mathcal{J}_{\text{GRPO}}(\theta) 
&=
\mathbb{E}_{q \sim P(Q), \{o_i\}_{i=1}^G \sim \pi_{\theta_{\text{old}}}(\cdot|q)}
\textcolor{red}{\frac{1}{\sum_{i=1}^G |o_i|}}\sum_{i=1}^{G} \sum_{t=1}^{|o_i|} \notag \\ 
& \min\left[\frac{\pi_\theta(o_{i,t}|q, o_{i,<t})}{\pi_{\theta_{\text{old}}}(o_{i,t}|q, o_{i,<t})} \hat{A}_{i,t}^{\textcolor{red}{\text{norm}}}, \; \text{clip}(\frac{\pi_\theta(o_{i,t}|q, o_{i,<t})}{\pi_{\theta_{\text{old}}}(o_{i,t}|q, o_{i,<t})}, 1-\textcolor{red}{\varepsilon_{\text{low}}}, 1+\textcolor{red}{\varepsilon_{\text{high}}})\hat{A}_{i,t}^{\textcolor{red}{\text{norm}}}\right], \notag \\
& \textcolor{red}{{\text{s. t. } \exists\ 1 \leq m < n \leq G, r_m \neq r_n}.}
\end{align*}

\subsection{Reward shaping}
\label{subsec:reward}

\looseness=-1 Choosing the appropriate reward is crucial for the RL algorithm to work effectively.
During training, model generations are evaluated along four axes: formatting, correctness, length, and language consistency, which we describe below.

\subsubsection{Formatting}

For both math and code problems, we instruct the model to follow a specific format, which facilitates the extraction of the model's answer:

\begin{enumerate}
    \item \textbf{Tag requirements:} (i) The model response must start with a \texttt{<think>} tag and must include a corresponding \texttt{</think>} tag. (ii) There should be exactly one set of these tags present in the response.

    \item \textbf{Mathematical responses:} For mathematical outputs, the response must include the final answer enclosed in \texttt{\textbackslash boxed\{\}} within the answer section, following the \texttt{</think>} tag.

    \item \textbf{Code responses:} For code outputs, the response must include at least one markdown block, formatted with triple backticks followed by the programming language specification, in the answer section.
\end{enumerate}

Failure to meet any of these conditions results in a reward of 0, and the response will not be graded further. Otherwise, the response gets a reward of 0.1 and proceeds to grading.

\subsubsection{Correctness}
\label{subsec:correctness}

If the generated answer follows the required formatting, we extract the model solution and use a verifier to assess its correctness.

\looseness=-1 \textbf{Math correctness.} The final answer is extracted from inside the last \texttt{\textbackslash boxed\{\}} in the solution and compared against the reference answer using a rule-based verifier. It normalizes both the ground-truth and the generated answer to correctly reward semantically identical responses with different syntaxes. 
We leverage a combination of different parsers and SymPy\footnote{https://www.sympy.org/en/index.html} to evaluate outputs and compare them to the original ground truth. 
An additional reward of 0.9 is given if the answer is correct, making the total reward 1.0.

\textbf{Code correctness.} 
Code is extracted from the first markdown code block in the answer section. 
If the code is written in C++, it is compiled with a timeout of 10 seconds, using the C++20 standard.
We pre-compile the \texttt{bits/stdc++.h} standard library header, which is commonly used in competitive programming, to speed up the compilation process.
We randomly select 20 tests from the available test cases, ensuring that the same tests are used within a given response group.
The code is then executed against these tests, with each test having a timeout of 4 seconds and a memory limit of 300 MB.
An additional reward of 0.9 is given if the code successfully passes all the tests.

\subsubsection{Length penalty}
\label{subsec:length-penalty}

Following~\citep{yu2025dapo}, we use soft length penalty to signal the model that the hard cutoff on maximal completion length is near. We fix two lengths $l_{\text{max}}$ and $l_{\text{cache}}$ and compute length penalty as
\begin{equation}
    R_{\text{length}}(y) = \begin{cases}
        0, & |y| \leq l_{\text{max}} - l_{\text{cache}} \\
        -0.1\cdot\frac{|y| - l_{\text{max}} + l_{\text{cache}}}{l_{\text{cache}}},  &l_{\text{max}} - l_{\text{cache}} < |y| \leq l_{\text{max}}, \\
        -0.1, & l_{\text{max}} < |y|
    \end{cases}
\end{equation}

\subsubsection{Language consistency reward}

A core design principle for Magistral is for it to reason in the same language as the user.
Reinforcement learning on math and coding problems without any treatment often results in mixed-language model responses.
In preliminary experiments without language constraints, we frequently observed outputs that mixed English, Chinese, and Russian words.
While these outputs were coherent, they were undesirable from a user perspective.

To prevent language switching, we 
translated 10\% of our problems written in English to the following languages: French, Spanish, Italian, German, Chinese, and Russian.
When calculating the reward for a conversation---a triple of $($problem, thoughts, answer$)$---we first normalized each of the three components by removing LaTeX content and code blocks, and then applied a fastText classifier~\citep{joulin2016fasttext} to each. If the classifier indicates that all three parts used the same language, we give an additional reward of $0.1$.

These simple modifications are sufficient to enable the model to closely follow the language of the user, with minimal code-switching, while maintaining performance on reasoning tasks.
Although we only translated the original English problems into a few languages, we observed that the model could successfully generate chains of thought in arbitrary languages.

\textbf{System prompt.}
We specify the format and the language requirements in the system prompt, which can be found in Figure~\ref{fig:task_solving_template}.
We find that RL training is quite sensitive to the system prompt we use. For example, the \greyblockmono{\small \textbf{Be as casual
and as long as you want}} part of the system prompt increases the entropy of the model and therefore improves the exploration of the model.

\begin{figure}[t]
\centering
\begin{systempromptbox}
A user will ask you to solve a task. You should first draft your thinking process (inner monologue) until you have derived the final answer. Afterwards, write a self-contained summary of your thoughts (i.e. your summary should be succinct but contain all the critical steps you needed to reach the conclusion). You should use Markdown and Latex to format your response. Write both your thoughts and summary in the same language as the task posed by the user.\\

Your thinking process must follow the template below:

<think>

Your thoughts or/and draft, like working through an exercise on scratch paper. Be as casual and as long as you want until you are confident to generate a correct answer.

</think>\\

Here, provide a concise summary that reflects your reasoning and presents a clear final answer to the user.\\

Problem:\\

\{problem\}
\end{systempromptbox}
\caption{
\textbf{Magistral's system prompt.}
The system prompt spells out the format and language guidelines for the model.
The same system prompt is utilized for both mathematical and coding problems.
\label{fig:task_solving_template}
}
\vspace{-0.2in}

\end{figure}

\section{Infrastructure}
\label{sec:infra}

\begin{figure}[t]
\centering
\includegraphics[width=1.0\textwidth]{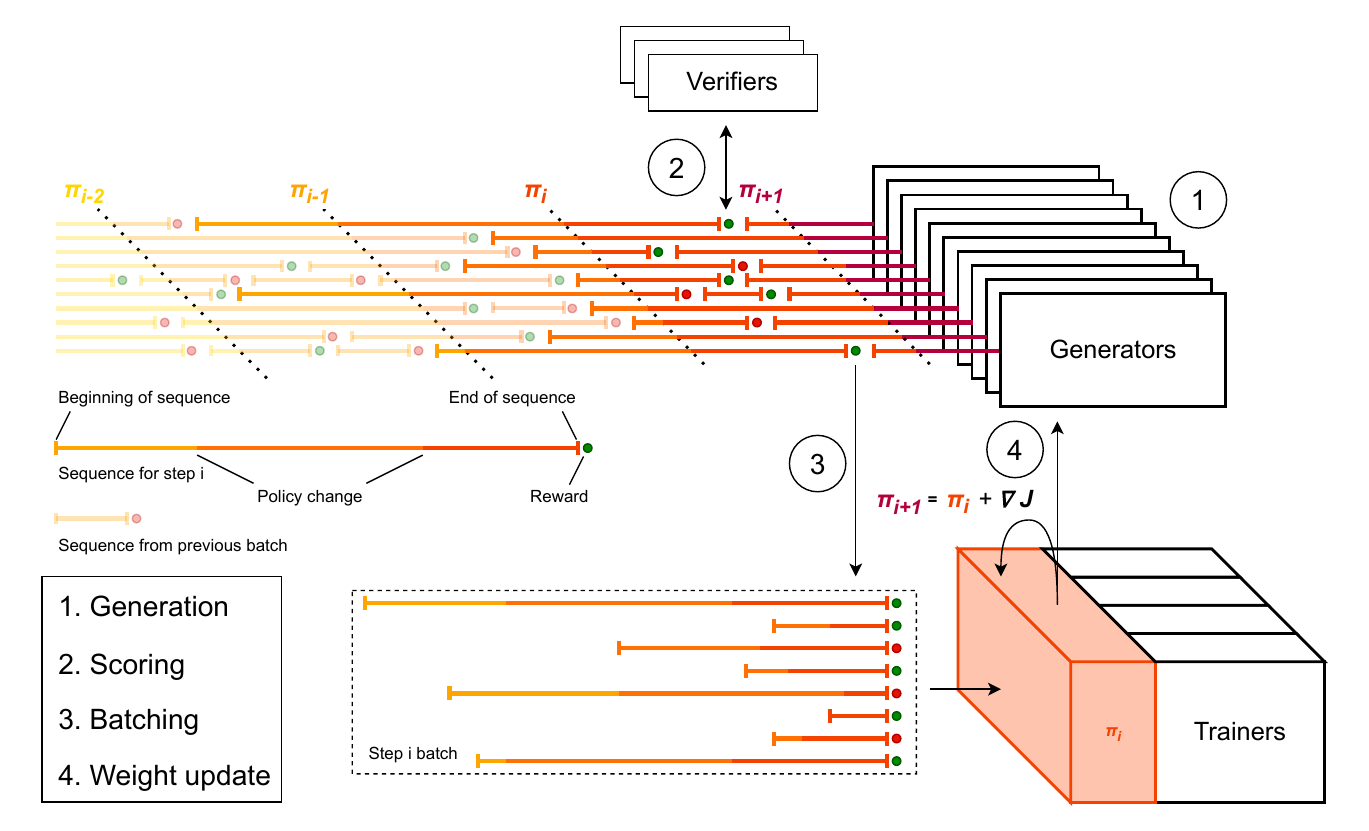}
\caption{
\looseness=-1 \textbf{Online training pipeline.}
1) Generators continuously output completions to prompts from input data sources. 2) Whenever a completion is finished, it is sent to the appropriate verifier. 3) Each sequence is sent to a different data parallel group using a pre-set permutation until every data parallel group has enough sequences to form a batch. 4) A single gradient step is performed and the trainer and generators are updated. In the generators, weights are replaced mid-generation, which means that in-flight generations continue with a slightly outdated key-value cache, as we do not refresh the cache.
Since the model resides on GPUs in both the trainer and the generators, the weights are transferred using NCCL for optimal performance. The model weights are dynamically consolidated to accommodate the different sharding topologies between trainers and generators.
\label{fig:training_pipeline}
}
\end{figure}

In this section, we present our infrastructure for online training. We adopt a distributed RL training system similar to those proposed in several prior works \citep{espeholt2018impala,hu2024openrlhf,noukhovitch2024asynchronous,sheng2024hybridflow,wu2025llamarl} that coordinates three kinds of workers:

\begin{itemize}
\item \textbf{Trainers} maintain the main copy of the model weights and perform gradient updates.
\item \textbf{Generators} perform `roll-outs', using the latest policy to return completions with log-probabilities from the training prompts.
\item \textbf{Verifiers} evaluate the completions produced by the generators and return a reward (see Section~\ref{subsec:reward} for details).
\end{itemize}

\paragraph{Challenges with distributed RL.}
Generators are a significant part of the total compute and the part that's unique to online RL.
Their workload is highly heterogeneous and hard to predict as the distribution of sequence lengths is highly skewed and changes over the course of training: the longest completions can take up to 5 times longer than the shortest.
One of the main constraints of the system is to introduce no bias on sequence lengths: the distribution of completion lengths must be exactly that of the training data, even though shorter completions finish more quickly.
A competing goal is to update the generator weights as soon as possible. We want the generations to be as on-policy as possible, but we also want the generators to operate without waiting for each other or the trainers.

\looseness=-1 
\paragraph{Asynchronous generations.}
In order to train without any approximation, we could process batches sequentially: start generators on a batch, wait for all sequences to complete, update the model weights for both trainers and generators, and repeat. However, this approach leads to idle generators and low pipeline efficiency due to heterogeneous completion times.
Instead, we prioritize efficiency and operate the generators continuously at maximum throughput without ever waiting for the trainers. We constantly gather groups from the generators, verify them, and update the trainers. After these updates, the trainers send the new weights to the generators via NCCL, without discarding the in-flight sequences currently being generated. Broadcasting weights from GPUs to GPUs is crucial as it reduces the time required for a single update to below 5 seconds, even with large models and large world sizes. We illustrate this process in Figure~\ref{fig:training_pipeline}.

As a solution is generated for a single prompt, it may experience multiple updates to the model weights, reflecting the latest improvements from the trainers. By the time it is fully processed and sent to the trainers, the model weights may have been updated several times, but the latest tokens are always generated on-policy.
When updating the model weights, the hidden states previously stored in the key-value cache become slightly outdated because they were computed by previous versions of the model. For performance, we find that recomputing the key-value cache is not necessary, potentially due to off-policy corrections inherent to the loss function \citep{schulman2017proximal}.

\looseness=-1 
\paragraph{Trainer optimization.}
We define a batch as a fixed number of generated completions, rather than a fixed number of tokens. Generators send each finished completion to a random trainer rank according to a pre-set permutation. A gradient update is performed when each data parallel rank has received enough completions to make a batch. If the trainers are the bottleneck, as is the case in early training when the generations are still short, we accumulate incoming generations into a blocking queue with a fixed size limit that controls off-policy degree. A batch may be partitioned into minibatches to perform several optimization steps (see Section~\ref{sec:batch-size}). Each minibatch has a fixed number of completions but a variable number of tokens, so it is further divided into microbatches of a fixed token size. Since we accumulate the gradient over microbatches, the order of samples does not matter. We take advantage of this property to implement a greedy collation algorithm, sorting the sequences by descending size and trying to fit them one by one into a free microbatch if there is one or starting a new otherwise. This ensures a homogeneous workload across training workers for each minibatch, reducing padding by 19\%.

\section{Data curation}
\label{sec: data}

We limit ourselves to problems with verifiable solutions; we use mathematical problems whose solution is a numerical answer or expression, and code problems with associated tests. 
We apply extensive filtering, which we describe here.

\subsection{Math}

\paragraph{Format filtering.}
\looseness=-1 We started with a large but noisy problem set of around 700k samples.
We first perform comprehensive pre-processing and filtering of the data to ensure all the problems are complete and that the final answers were accurate and verifiable with a rule-based system.
Particularly, we filter proof-based and multi-part problems for which it is difficult to verify correctness.
Furthermore, we reformulate multiple-choice problems into statement-based problems for more robust verification and increased difficulty.

\paragraph{Difficulty filtering.}
We implemented a two-stage filtering pipeline to curate a dataset of problems at a `goldilocks' difficulty level, neither too easy nor too hard for the model to learn from.  First, we performed an initial difficulty assessment using Mistral~Large~2~\citep{mistrallarge2}, by sampling $16$ solutions for each problem and removing the ones that are either never solved or solved with a high success rate. This initial, curated set of problems was then used to train a 24B model via our online RL pipeline, resulting in a small but capable checkpoint which we use solely for grading. 

In the second stage, this stronger, RL-trained model was used to re-grade the entire original dataset. We again sampled 16 responses for each problem, filtering out the easiest and the still-unsolved problems. We then further filter out potentially incorrect problems where a majority of samples have the same final answer but disagree with the ``ground-truth’’ answer. This is because when the model consistently reaches a consensus that contradicts the reference solution, the problems themselves are more likely to have wrong ground-truth answers.

This two-stage methodology was crucial because a single pass with the initial, weaker Mistral~Large~2 model would have been insufficient. Its reasoning capabilities would likely have caused it to discard many genuinely difficult problems by incorrectly classifying them as unsolvable. By using a stronger, RL-trained model for a second pass, we ensured that these valuable and challenging training examples were accurately assessed and retained.

\begin{table}[h]
    \centering
    \vspace{-0.2cm}
    \caption{
    Number of math training samples after different filtering stages.
    \label{tab:mathdata}
    }
    \vspace{0.2cm}
    \begin{tabular}{cccc}
    \toprule
     Initial data& w/ Format filtering & w/ Difficulty filtering \\
    \midrule
     699k & 501k & 38k \\
    \bottomrule
    \end{tabular}
    \vspace{-0.3cm}
\end{table}

\subsection{Code}

We gathered code contest data from various sources. Each data point includes a problem statement and, when available, correct solutions and related tests. For the training process, we want problem statements and a large number of correct tests per problem. In order to achieve this, we first remove any problems without solutions and without enough tests. 
Each solution is then executed on all available tests, and we discard tests with insufficient agreement. For tests with sufficient agreement but where no solution succeeded, we assume that the test is incorrect and update it to reflect the most common result among the solutions' outputs.
In cases where code problems lack tests, we generate additional tests and subject them to the same evaluation process. 

\looseness=-1 Finally, where applicable, problem statements are duplicated to require code in Python or C++, two commonly used languages in competitive programming. This process resulted in a dataset of 35k code problems.

\section{Experiment and results}
\label{sec:results}

\begin{figure}[t]
    \centering
\includegraphics[width=1.0\textwidth, trim = 0cm 16cm 0cm 6cm, clip]{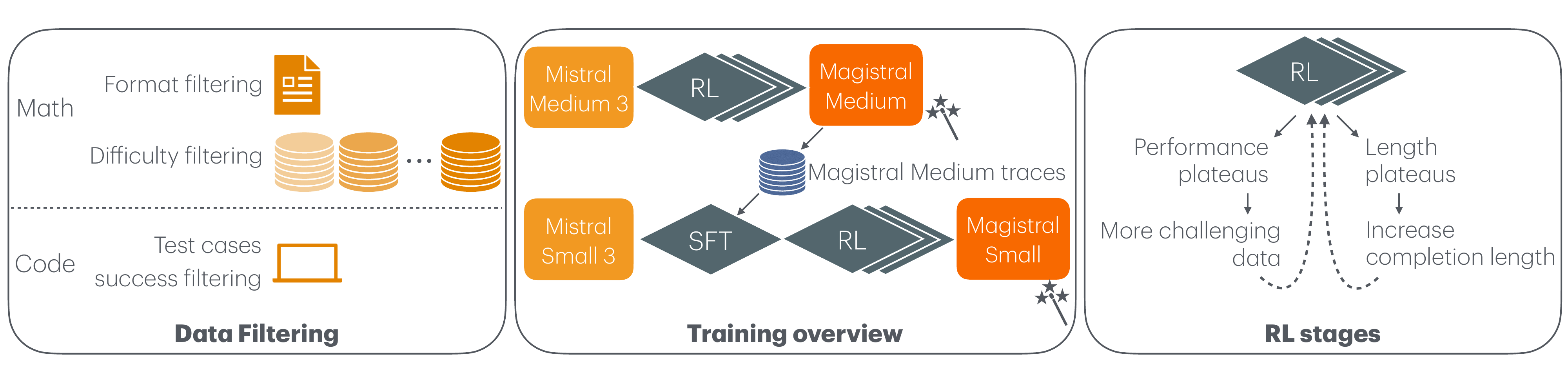}
    \caption{
    \looseness=-1 \textbf{Overview of the filtering, training and RL stages discussed in the paper}. We do RL over Mistral~Medium~3 to get Magistral Medium. We use this model to generate answers for a large set of diverse prompts. We use these generated traces to finetune Mistral~Small~3 and then perform RL to get Magistral Small.
    \label{fig:relationship}
    }
    \vspace{-0.1in}
\end{figure}

In this section we present the Magistral models. 
Our goal is to answer two questions: 
(i) how far can one get with pure reinforcement learning on a large base model? 
(ii) given a strong teacher model, how can one achieve the strongest possible lightweight model?
To this end, we trained Magistral Medium, on top of Mistral~Medium~3~\citep{mistralmedium3} with pure RL; and Magistral Small, which began with SFT traces derived from Magistral Medium.

\subsection{Evaluation benchmarks and baselines}

\looseness=-1 We report results on benchmarks that assess capabilities in the fields of mathematics, coding, and STEM. For math, we provide results on the American Invitational Mathematics Examination benchmarks (AIME'24, AIME'25), and on the MATH dataset~\citep{hendrycks2021measuring}. 
In coding, we include LiveCodeBench (both v5 and v6 versions)~\citep{jain2024livecodebench}, and Aider Polyglot~\citep{aider_polyglot_benchmark}. 
For STEM, we report results based on the GPQA dataset~\citep{rein2024gpqa}. Additionally, we also report our results on the text-only questions from Humanity's Last Exam~\citep{phan2025humanity} which comprises of 2,500 questions across dozens of subjects, including mathematics, humanities, and natural sciences. For all evaluation tasks, we set the temperature to 0.7 and use a top-p of 1.0 for Math evals and GPQA, and 0.95 for coding tasks. The maximum token length is set to 40k for AIME and LiveCodeBench, and 32k for all other evaluations.

For baselines we include results from~\citep{deepseekai2025deepseekr1}, which reports comparable datapoints for training with RL at scale, both with and without SFT on traces from a reasoning model.

\subsection{Magistral Medium -- reasoning RL from scratch}

Here our goal is to evaluate the quality of our RL stack by training a model without any `cold start' (i.e priming for reasoning by distillation of reasoning traces).
We used Mistral~Medium~3 Instruct~\citep{mistralmedium3} as the starting checkpoint for this run. Training was done in multiple stages with distinct hyper-parameters.
Particularly, the stages were designed to ensure the following criteria were always satisfied:
\begin{enumerate}
    \item \textbf{Dataset is not too easy.} As the model performance increases, we increase the difficulty of the data. 
    Harder data splits are constructed by including more complicated data (which were filtered out in earlier stages) or removing completely solved problems from the data. 
    \item \textbf{Generation length does not stop growing.} To prevent stagnation in generation length, 
    we increase both maximal allowed completion length and maximal completion length $l_{\text{max}} - l_{\text{cache}}$ not punished by length penalty (c.f.~Section~\ref{subsec:length-penalty}). 
    We increased $l_{\text{max}} - l_{\text{cache}}$ twice as $16\text{k}\to 24\text{k}$ and $24\text{k}\to 32\text{k}$.
    \item \textbf{KV-cache memory burden is not too large.} As generation length increases, the memory usage associated with the KV cache increases. 
    To address this, we scale down the total number of concurrent requests running $n_{\text{async}}$, the batch size $n_{\text{batch}}$, and the minibatch size $n_{\text{minibatch}}$. The impact of batch size is discussed in Section~\ref{sec:batch-size}. During training we decreased batch size twice as $8\text{k}\to 4\text{k}$ and $4\text{k}\to2\text{k}$.
\end{enumerate}

Table~\ref{tab:benchmark-results} shows the results of Magistral Medium trained with pure RL, compared against analogous experiments from~\citep{deepseekai2025deepseekr1}.
We find our RL pipeline alone yields a nearly 50\% accuracy increase in AIME '24 (pass@1), and 30\% on LiveCodeBench (v5).

\subsection{Magistral Small -- RL on top of reasoning SFT bootstrapping}
Given a strong `teacher' model in Magistral Medium, we next explore how one can train the strongest possible student model.
To do so, we train Magistral Small, which is `cold-started' with SFT traces from Magistral Medium.

In contrast with pure RL training (which benefits from a small set of extremely clean and difficult training points, Section~\ref{sec: data}), we find diversity of prompts to be important for the reasoning cold-start.
We begin by extracting traces with correct answers from the RL training of Magistral Medium, excluding those from early steps with short CoTs. We also maintain a mixed difficulty level of the problems by limiting number of generations per problem to avoid biasing the collected traces towards easier problems and also upsampling problems with lower pass rates.

We augment this SFT cold-start data by generating responses from our Magistral Medium on a large set of diverse prompts, sourced from OpenThoughts~\citep{guha2025openthoughtsdatarecipesreasoning} and the code subset of OpenR1~\citep{openr1, penedo2025codeforces}. We perform additional filtering on top and kept a subset of the prompts.
This gives us a reasoning dataset with mixed difficulty. We also include 10\% of datapoints for general instruction tuning in order to preserve non-reasoning capabilities.
We finetuned Mistral~Small~3 Instruct (a 24-billion parameter model) for 4 epochs, and chose the best checkpoint on AIME'24 as the initial checkpoint for the following RL stage.

We then trained this SFT checkpoint with RL using a batch size of 2048 sequences, and a maximum non-penalized completion length $l_{\text{max}} - l_{\text{cache}}$ of 32k. We used a sampling temperature of 1.0 for our generations, as it provided the best balance between avoiding the lack of diversity seen at lower temperatures and the incoherent outputs generated at higher temperatures. We use a $\varepsilon_{\text{high}}$ of 0.3, to encourage exploration, as the cold-started model yielded responses with far lower entropy.  

Table~\ref{tab:24b-results} shows the performance of the 24B model trained under three different paradigms: with SFT 
alone; with RL alone; and with RL on top of the cold-start checkpoint.
Here, contrary to findings from~\citep{deepseekai2025deepseekr1}, we find one can get substantial boosts with RL even on a smaller base model, over and above distillation from the larger teacher.
This underscores the strength of the RL stack introduced in this work.

\begin{table}[!htbp]
\centering
\caption{\textbf{Results of Magistral Medium trained solely with RL.} To reduce variance, we compute the average over 64 runs for AIME (shown as pass@1/maj@64) and over 16 runs for LiveCodeBench. Humanity's Last Exam is evaluated only for the text subset.}
\label{tab:benchmark-results}
\vspace{0.2cm}
\renewcommand{\arraystretch}{1.3}

\begin{tabularx}{\textwidth}{lccccc}
\toprule
\textbf{Task} & \multicolumn{1}{X}{\textbf{Mistral Medium 3}} & \multicolumn{1}{X}{\textbf{Magistral Medium}} & \multicolumn{1}{X}{\textbf{DeepSeek-v3}} & \multicolumn{1}{X}{\textbf{DeepSeek-R1-Zero}} & \multicolumn{1}{X}{\textbf{DeepSeek-R1}}\\
\midrule
Reasoning SFT before RL & - & \xmark & - & \xmark & \cmark\\
\midrule
AIME'24                          & 26.8 / 43.4 & 73.6 / 90.0 & 39.2 & 71.0 & 79.8\\
AIME'25                          & 21.2 / 30.0 & 64.9 / 83.3 & 28.8 & - & 70.0\\ 
MATH-500                         & 91.0 & 94.3 & 90.2 & 95.9 & 97.3\\
GPQA                             & 59.6 & 70.8 & 59.1 & 73.3 & 71.5\\
LiveCodeBench (v5)                & 29.1 & 59.4 & 36.2 & 50.0 & 65.9\\
Aider Polyglot                   & 28.9 & 47.1 & 49.6 & - & 53.3\\
LiveCodeBench (v6)                & 30.0 & 50.3 & - & - & -\\
Humanity's Last Exam      & 4.4&  9.0 & - & - & 8.6\\
\bottomrule
\end{tabularx}
\end{table} 

\begin{table}[!htbp]
\centering
\caption{\textbf{Performance of Magistral Small compared with different training setups across various benchmarks.} We report the performance of three distinct 24B models: Mistral Small 24B fine-tuned on reasoning traces from Magistral Medium (SFT), Mistral Small 24B trained from scratch with RL (RL only), and Mistral Small 24B fine-tuned on Magistral Medium traces and subsequently enhanced with RL (SFT + RL) which is the final Magistral Small.
We observe that the combination of fine-tuning on reasoning traces with RL leads to the best performance.
For the evaluation of Humanity's Last Exam, only the text subset was considered. \label{tab:24b-results}}
\vspace{0.2cm}
\renewcommand{\arraystretch}{1.3}
\begin{tabular}{lccc}
\toprule
\textbf{Task} & \textbf{SFT} & \textbf{RL-only} & \textbf{SFT + RL (Magistral Small)} \\
\midrule
AIME'24$_{\text{pass@1}}$              & 65.4 & 65.8 & 70.7 \\
AIME'24$_{\text{maj@64}}$              & 90.0 & 86.7 & 83.3 \\
AIME'25$_{\text{pass@1}}$              & 55.6 & 51.9 & 62.8 \\
AIME'25$_{\text{maj@64}}$              & 76.7 & 66.7 & 76.7 \\
MATH-500                               & 93.2 & 95.4 & 95.9 \\
GPQA                                   & 63.4 & 68.8 & 68.2 \\
LiveCodeBench (v5)                     & 52.2 & 46.4 & 55.8 \\
LiveCodeBench (v6)                      & 44.6 & 42.4 & 47.4 \\
Humanity's Last Exam             & 5.3 & 6.1 & 6.4 \\
\bottomrule
\end{tabular}
\end{table}

\subsection{Multilingual benchmarks}

\looseness=-1 To evaluate Magistral's multilingual capabilities, we interacted with Magistral Medium in multiple languages to check that it could reason and answer in the user's language. We also tested Magistral Medium on multilingual~(French, Spanish, German, Italian, Russian, and Chinese) versions of the AIME 2024 benchmark. These multilingual versions were created by translating the questions from English into each of the languages. The results are presented in Table~\ref{tab:multilingual benchmarks}. We see that the model performs 4.3-9.9\% lower on multilingual versions compared to English, which corresponds to 1-3 questions on the actual AIME test, possibly because we constrained the language of reasoning. This degradation is roughly similar to that of the base model.
Note that on the multilingual benchmarks, all of the reasoning and the final response are conducted in the input language (i.e., not English).

\begin{table}[!htbp]
\centering
\caption{
Magistral Medium's pass@1 performance on multilingual versions of the AIME 2024 benchmark.
\label{tab:multilingual benchmarks}
}
\vspace{0.2cm}
\renewcommand{\arraystretch}{1.3}
\begin{tabularx}{\textwidth}{lccccccc}
\toprule
Language & English & French & Spanish & German & Italian & Russian & Chinese \\
\midrule
AIME'24 (pass@1) & 73.6
& 68.5
& 69.3
& 66.8
& 66.7
& 65.0
& 63.7 \\
\bottomrule
\end{tabularx}
\end{table}

\section{Ablations}
\label{sec:ablations}
In this section, we tweak parameters of the training process to investigate what happens when RL is performed on only one modality, compare RL to the distillation SFT baseline, and shed light on two training choices we had to reckon with, batch and minibatch size, and advantage normalization.

\subsection{Cross-domain generalization}
\label{sec:cross-domain-transfer}

\looseness=-1 We investigate the ability of our model to generalize across domains by training on one domain (math or code) and evaluating on the other. Specifically, we conduct two experiments on the 24B model: one where the model is trained exclusively on math data and evaluated on both math and code, and another where it is trained only on code and evaluated similarly. As shown in Table~\ref{tab:cross-domain generalization}, the model demonstrates strong performance to out-of-domain tasks, showcasing the generalization ability of RL.

\begin{table}[b]
\centering
\caption{
Cross-domain generalization during math-only and code-only RL for a 24B model
\label{tab:cross-domain generalization}
}
\vspace{0.2cm}
\renewcommand{\arraystretch}{1.3}
\begin{tabularx}{0.6\textwidth}{lcc}
\toprule
Model & AIME'24 & LiveCodeBench v5 \\
\midrule
Starting Checkpoint & 32.2 & 22.7 \\
RL (Math only) & 62.5 & 38.3 \small{(\textbf{+15.6})} \\
RL (Code only) & 49.7 \small{(\textbf{+17.5})} & 42.7 \\
\bottomrule
\end{tabularx}
\end{table}

\subsection{Distillation vs. RL for small models}
\label{sec:distillation-small-models}
\begin{figure}[h]{
    \centering
    \includegraphics[width=\textwidth]{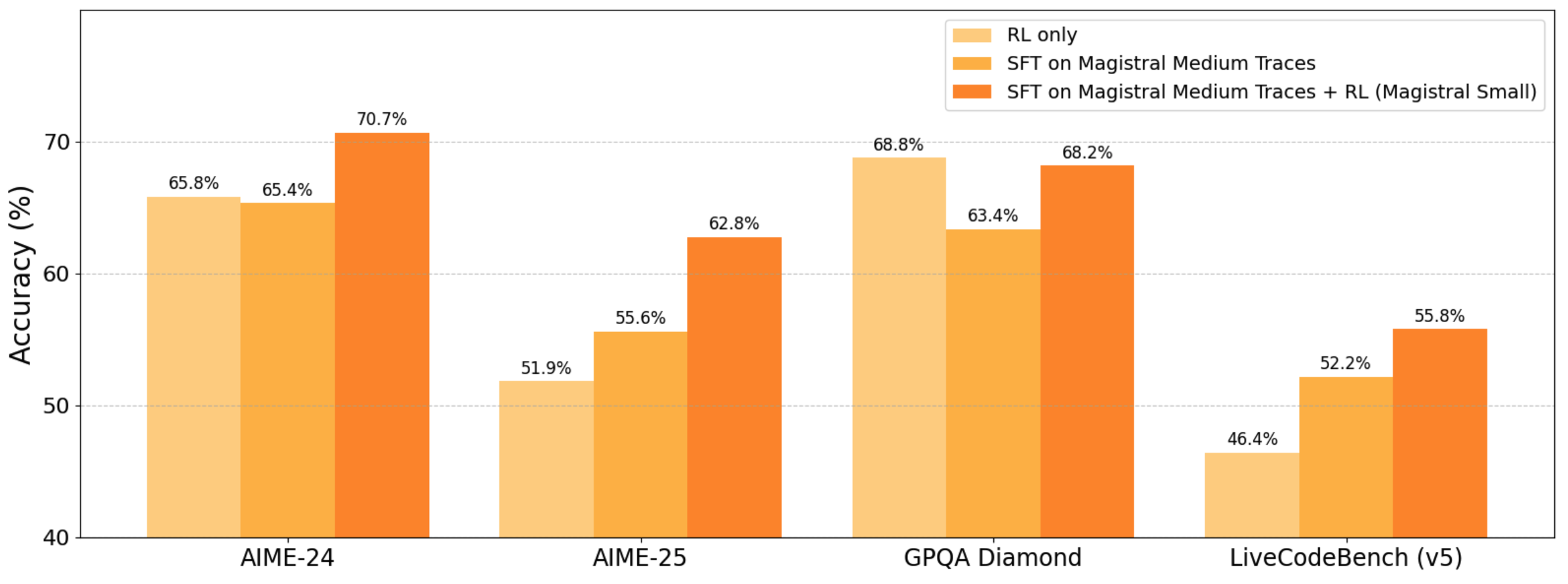}
    \caption{
    \looseness=-1
    \textbf{Performance of Magistral Small compared with different training setups on various benchmarks.}
    We report the performance of three distinct 24B models: Mistral Small 24B trained from scratch with RL (RL only), Mistral Small 24B fine-tuned on reasoning traces from Magistral Medium, and Mistral Small 24B fine-tuned on Magistral Medium traces and subsequently enhanced with RL, which is the final Magistral Small.
    We observe that the combination of fine-tuning on reasoning traces with RL leads to the best performance.
    }
    \label{fig:small_perf}
    }
\end{figure}

Previous works \citep{deepseekai2025deepseekr1} have observed that smaller models relying solely on RL may not be able to achieve performance comparable to those distilled from larger reasoning models. However, our findings contradict this observation: 
we achieved strong results even with pure RL on top of Mistral~Small~3. 
As shown in Figure~\ref{fig:small_perf}, our Mistral~Small~3 with pure RL achieves similar performance on AIME'24 as the distilled version. It even outperforms the distilled version on MATH and GPQA, but has slightly lower performance on code benchmarks such as LiveCodeBench. These results suggest that the benefits of RL are not exclusive to larger base models and hold equally well for smaller models.  Furthermore, our findings indicate that the RL on top of the distilled checkpoint can yield even better performance, leading to over 5 points gain across various benchmarks.

\subsection{Batch and minibatch size}
\label{sec:batch-size}

Reinforcement learning (RL) algorithms like PPO or GRPO introduce two distinct batch scales. The batch size, denoted as $n_{\text{batch}}$, refers to the number of sequences collected before updating the generator's weights. The minibatch size, $n_{\text{minibatch}}$, indicates the number of sequences used to compute the gradient and perform a single optimization step. It is important to note that $n_{\text{minibatch}}$ must divide $n_{\text{batch}}$. Additionally, in an asynchronous RL pipeline, a third scale is introduced: the number of concurrent sequences, $n_{\text{async}}$, which represents the number of sequences being generated in parallel. If the number of concurrently generated sequences $n_{\text{async}}$ is much larger than the batch size $n_{\text{batch}}$, a typical sequence was generated with $n_{\text{async}} / n_{\text{batch}}$ different policies and could be too off-policy. The effect becomes worse as we do more than one minibatch update per one batch. 

To test this hypothesis we prepared a strong 3B model using SFT starting from Ministral 3B, and then trained it using GRPO on math-only data with a constant learning rate, a fixed $n_{\text{async}} = 4096$, and different values of $n_{\text{batch}}$ and $ n_{\text{minibatch}}$ in $\{1024, 2048, 4096, 8192\}$.

\begin{figure}[t]
    \centering
    \includegraphics[width=\linewidth]{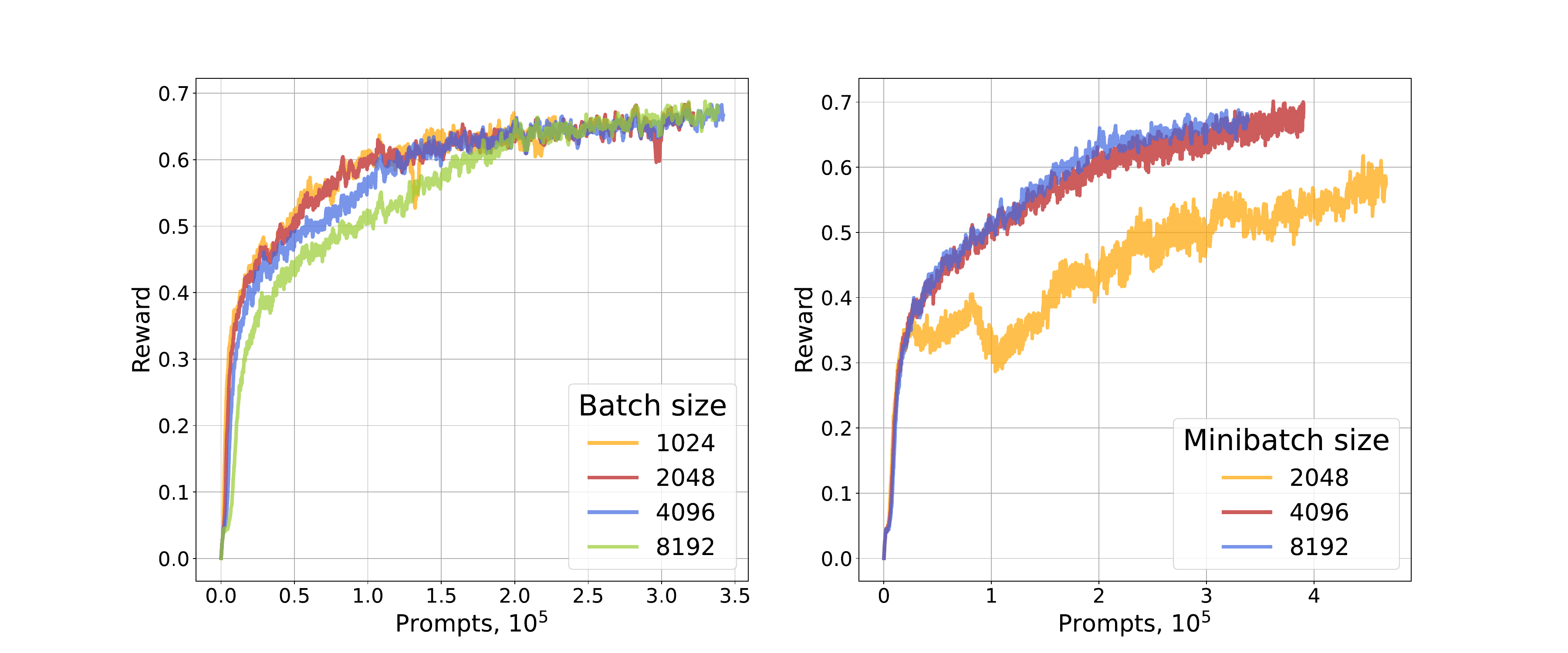}
    \caption{
    \looseness=-1
    \textbf{Impact of batch and minibatch sizes on RL training rewards.}
    (a)~Reward during RL training of 3B model on math data for different batch sizes, while keeping minibatch size equal to batch size. Number of concurrently generated sequences is kept constant at 4096.
    (b)~Reward during RL training in the same setup for different minibatch sizes at fixed batch size of 8192 sequences. We observe that performance doesn't depend strongly on batch size, but degrades when there are more than 2 minibatches in a batch.
    \label{fig:batch-size}
    }
\end{figure}

We observe that as long as we keep $n_{\text{batch}} = n_{\text{minibatch}}$ and that $n_{\text{batch}}$ is large enough, the performance is very similar when plotted depending on the number of processed prompts, as can be seen in Figure~\ref{fig:batch-size}~(a).
On the other hand, when $n_{\text{minibatch}}$ is decreased while keeping $n_{\text{batch}}$ constant, the performance suddenly degrades, even when compared to $n_{\text{batch}}$ reduced to the same $n_{\text{minibatch}}$, as highlighted in Figure~\ref{fig:batch-size}~(b).
When $n_{\text{batch}} \leq 1024$, the training becomes less stable, so we opt to keep ratio $n_{\text{async}} / n_{\text{batch}} \leq 2$ and $n_{\text{batch}} = n_{\text{minibatch}}$ during final training and further ablations.

\subsection{Advantage normalization}
\begin{figure}[H]
    \centering
    \includegraphics[width=1.0\textwidth]{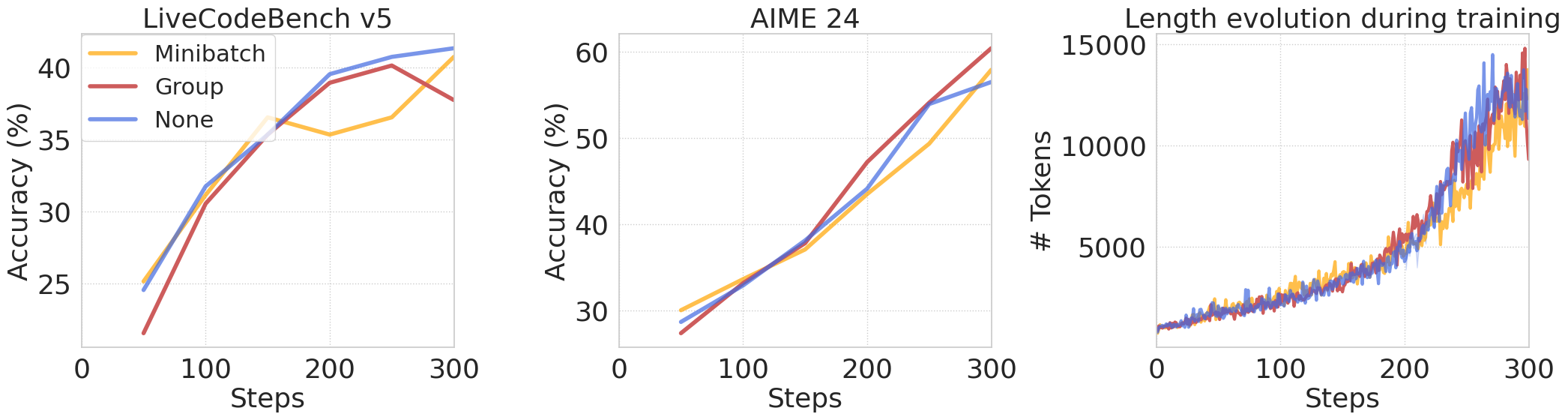}
    \caption{
    \label{fig:norm-ablation}
    \looseness=-1
    \textbf{Results for training with different advantage normalizations in GRPO.}
    We observe that different normalization methods do not lead to significant difference either in evaluation performance or the length growth during training.
    }
\end{figure}

We experimented with the following advantage normalization methods: 
\begin{itemize}\setlength\itemsep{0.1em}
    \item Minibatch - normalize advantages within a minibatch
    \item Group normalization - normalize advantages within a group over a single prompt
    \item No normalization - do not normalize advantages
\end{itemize}
 Previous works \citep{liu2025understanding, andrychowicz2020} have noted that normalization over a group of generations for a given question can lead to a bias where easy questions or hard questions are upweighted due to their lower standard deviation values. However, we did not observe any significant effects on evaluation performance or the growth of the length as shown in Figure~\ref{fig:norm-ablation}. Hence, we decided to use minibatch normalization for all our experiments. 
\label{sec:norm-ablation}

\section{Analysis}
\label{sec:analysis}

In this section, we investigate the dynamics of RL training and present evidence that increasing completion length is the main resource that improves the performance of the model. Those dynamics are not destructive to previous capabilities, and the reasoning capabilities can even generalize: multimodal reasoning gets improved for free, and function calling and instruction following remain unchanged or even get a small boost. Additionally, we discuss two ideas that didn't work for us - giving more fine-grained rewards in code tasks based on test completion rate and controlling entropy via entropy bonus term in the loss.  

\subsection{Reinforcement learning moves weights in low-dimensional space}

To better understand the dynamics of Magistral during RL training, we follow the method of ~\citep{li2018visualizinglosslandscapeneural} to analyze the \textbf{Magistral Small
RL-only} run
and visualize the loss landscape around the final checkpoint.

First, we stack the weights of all intermediate checkpoints in a matrix $X \in \mathbb{R}^{T\times W}$, where $T$ is the number of checkpoints and $W$ is the number of weights.
Then, we subtract the mean weights across the $T$ checkpoints and perform a PCA analysis to find two principal components of the matrix $X$ in the weight space.
Since the weight space is very high-dimensional, we use the iterative Lanczos-Arnoldi algorithm~\citep{saad2003iterative} to find the top-2 eigenvectors of $X^TX$.
As a result, we obtain two components $c_1$ and $c_2$ that we L2-normalize to have a unit norm.

Second, we perturb the final checkpoint weights $w^* \in \mathbb{R}^{W}$ by adding two components as 
\begin{equation}
    \label{eq:checkpoint-perturbation}
    w(\alpha_1, \alpha_2) = w^* + \alpha_1 c_1 + \alpha_2 c_2
\end{equation}
We evaluate each perturbed checkpoint on a fixed batch of 512 prompts, generating 16 completions per prompt, and using the same reward setting as in \textbf{Magistral Small
RL-only} run. Finally, we compute mean reward and mean output length for each checkpoint and plot it in $(\alpha_1, \alpha_2)$ coordinates.

\begin{figure}[!htbp]
    \centering
    \includegraphics[width=\linewidth]{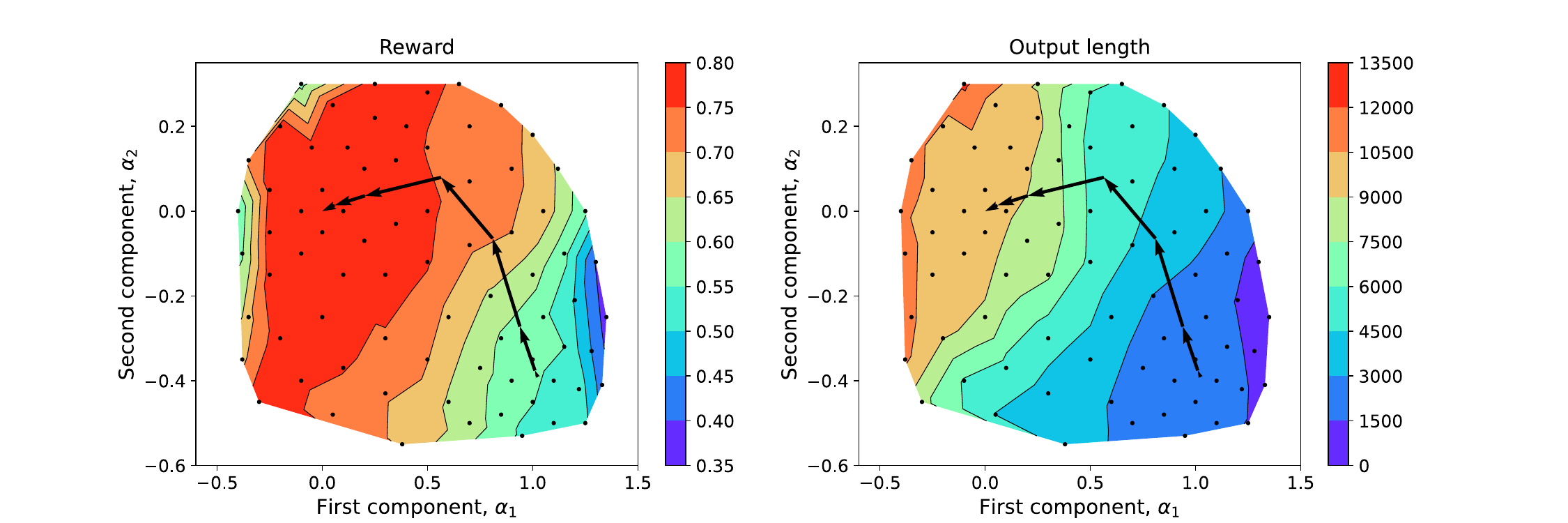}
    \caption{
    \looseness=-1
    \textbf{Reward and length evolution in $w(\alpha_1, \alpha_2)$ hyperplane.}
    Black arrow trajectory is a projection of intermediate checkpoints of \textbf{Magistral Small
RL-only} run on the hyperplane. Black points are perturbed checkpoints computed using Equation~\ref{eq:checkpoint-perturbation}. Intermediate values are computed with linear interpolation on the triangular grid.
    \label{fig:pca-evolution}
    }
\end{figure}

\begin{figure}[h]
    \centering
    \includegraphics[width=0.5\linewidth]{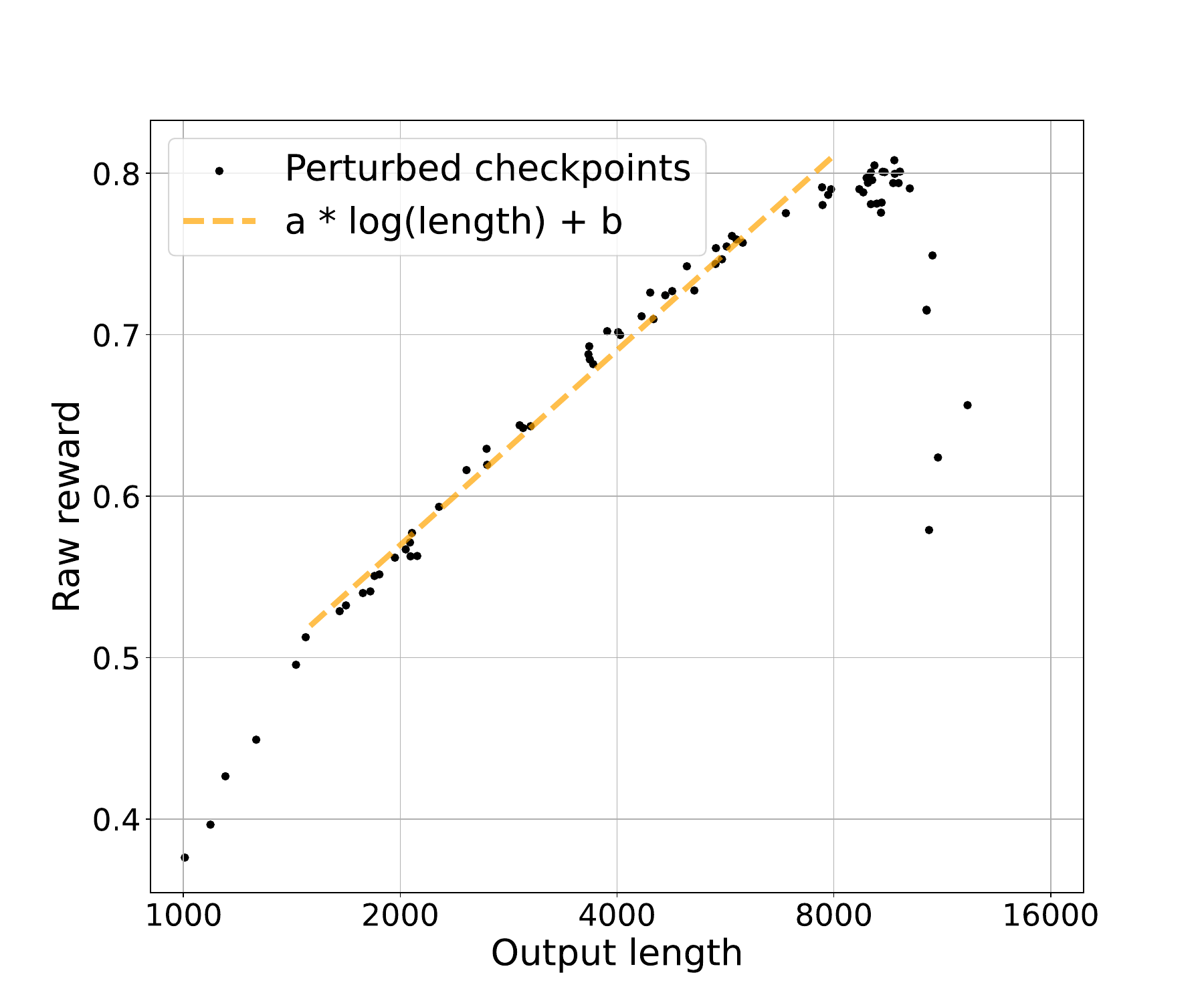}
    \caption{
    \textbf{Reward scaling with output length.}
    Each point corresponds to a perturbed checkpoint computed with Equation~\ref{eq:checkpoint-perturbation}. We generate 8192 completions with the checkpoint and evaluate mean output length and raw reward (reward without length penalty). We perform linear regression on checkpoints with mean output length between 1500 and 8000 and observe that reward scales logarithmically with the output length.
    \label{fig:length-scaling}
    }
\end{figure}

We clearly observe that there is a ``length'' direction - as model goes from right to left in Figure~\ref{fig:pca-evolution}, mean reward and output length grow up until the point where length starts to hit length penalty and maximally allowed completion length. We additionally plot dependence of raw reward without length penalty on output length, observing a ubiquitous log scaling in Figure~\ref{fig:length-scaling}.

\begin{figure}[h]
    \centering
\includegraphics[width=\textwidth]{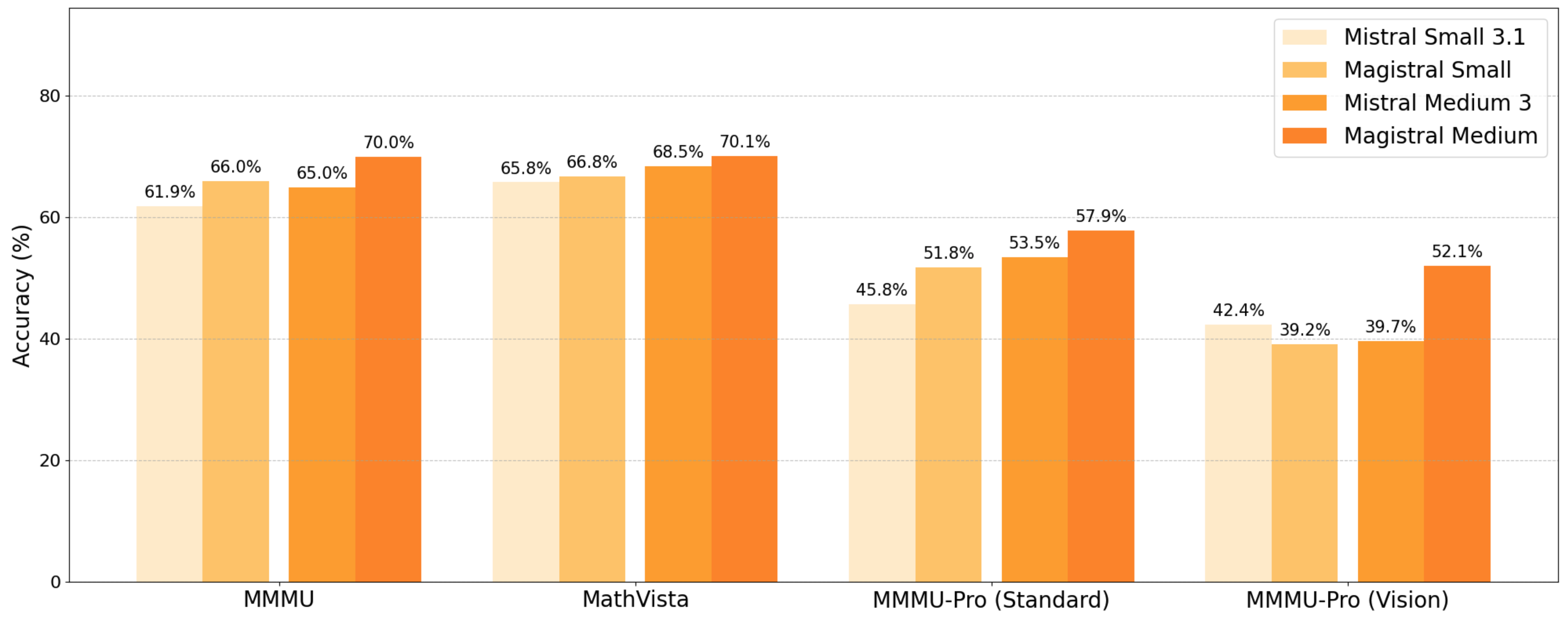}
    \caption{
    \textbf{Performance on multimodal benchmarks.}
    \label{fig:multimodal_perf}
    }
\end{figure}

\subsection{Eating the multimodal free lunch}
\label{sec:multimodal}

The initial checkpoints utilized for RL training, Mistral~Small~3 and Mistral~Medium~3, are multimodal models and come with associated vision encoders.
During the RL training phase, as the models are trained on text-only data, one might expect the multimodal performance to degrade.
However, on the contrary, we discover that the models not only retain their multimodal capabilities, but unexpectedly develop enhanced multimodal reasoning abilities.
The resulting models also showcase improved performance on multimodal benchmarks.

We report results multimodal benchmarks designed to assess reasoning capabilities, specifically MathVista~\citep{lu2024mathvista}, MMMU~\citep{yue2023mmmu}, and MMMU-Pro~\citep{yue2025mmmu-pro}.
Our results in Figure~\ref{fig:multimodal_perf} show no performance regression across most benchmarks, with notable improvements observed on MMMU (+5\%, reaching 70\%), MMMU-Pro-Standard (+4.4\%, reaching 57.9\%) and MMMU-Pro-Vision (+12\%, reaching 52.1\%). While the most significant improvements are seen in scientific questions that require textual reasoning, we observe that the model transfers its extended thinking process across all types of questions (see Figures~\ref{fig:mm-physics}~\ref{fig:mm-chemistry}~\ref{fig:mm-biology} for qualitative examples).

\subsection{Impact of RL on other capabilities}

Similar to the multimodal capabilities mentioned in Section~\ref{sec:multimodal},
our RL checkpoint maintains and even improves its tool calling and instruction following capabilities \citep{zhou2023instructionfollowingevaluationlargelanguage} (Table~\ref{tab:if_and_tc_vs_mm}).  This allows us to integrate the model out-of-the-box with existing tools as shown.

\begin{table}[h!]
\centering
\caption{
\textbf{Benchmarks before and after reinforcement learning.}
Internal bench is Mistral's internal function calling benchmark.
We use an internal version of IFEval that fixes some issues with the public version. The scores are not comparable with other publicly shared scores.}
\label{tab:if_and_tc_vs_mm}
\vspace{0.2cm}
\renewcommand{\arraystretch}{1.5}
\begin{tabularx}{0.9\textwidth}{llcc}
\toprule
\textbf{Category} & \textbf{Benchmark} & \textbf{Mistral~Medium~3} & \textbf{Magistral Medium} \\
\midrule
Function calling & Internal bench & 87.2 & 87.4 \\
Instruction following & IFEval & 86.8 & 87.4 \\

\bottomrule
\end{tabularx}
\end{table}

\subsection{Unsuccessful approaches}

In this section, we present various approaches that we tried but ultimately did not adopt, as they did not yield any performance improvements.

\subsubsection{Partial reward for code data}
\label{sec:partial-rewards}
The strict requirements of competitive programming, in terms of correctness and adherence to complexity constraints, result in sparse rewards, often causing many code generations to be discarded due to limited reward diversity.

\begin{figure}[htbp]
    \centering
    \begin{minipage}[t]{0.48\textwidth}
        \centering
        \vspace{0pt}
        \includegraphics[width=\linewidth]{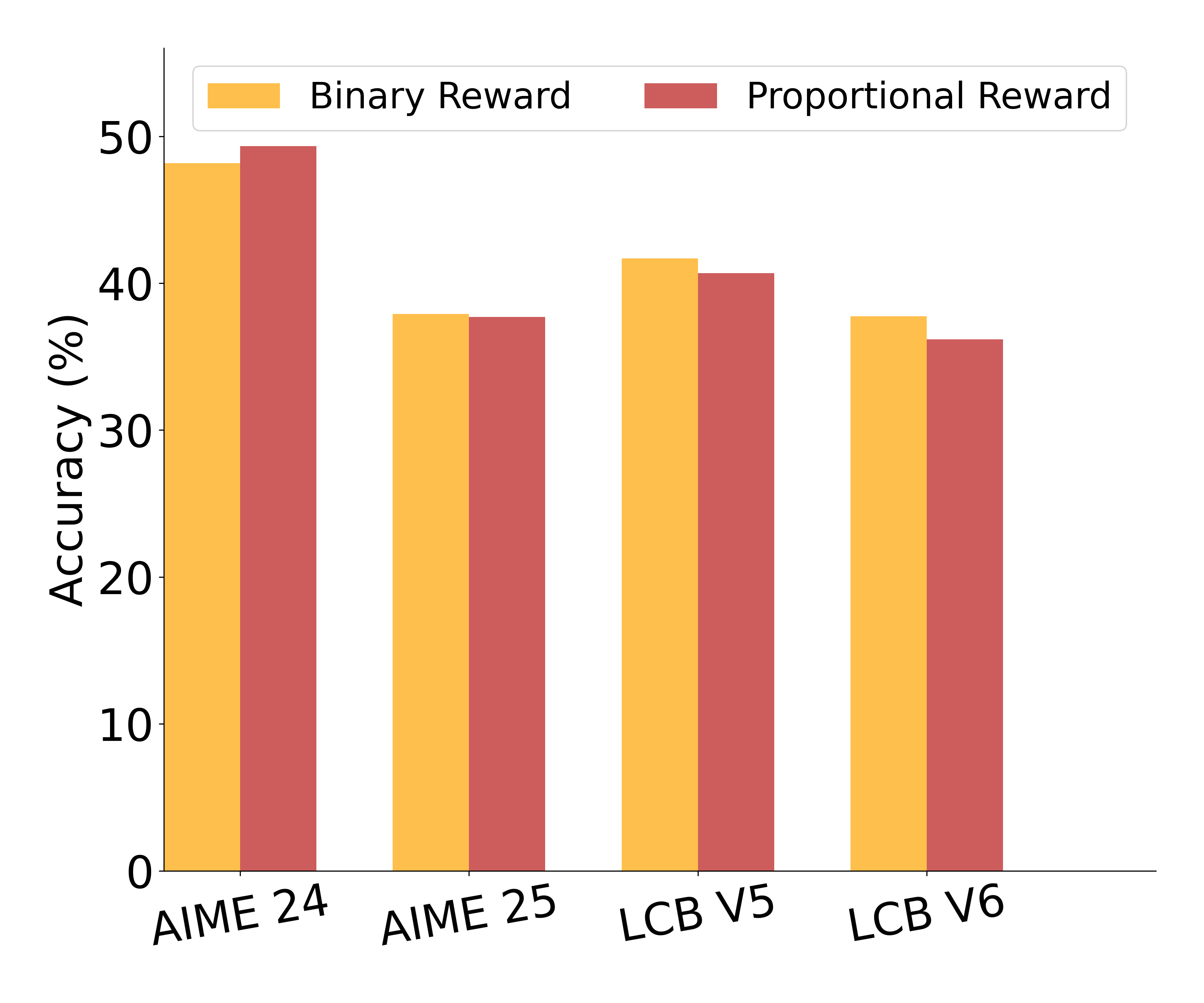}
    \end{minipage}\hfill
    \begin{minipage}[t]{0.48\textwidth}
        \centering
        \vspace{3pt}
        \includegraphics[width=\linewidth]{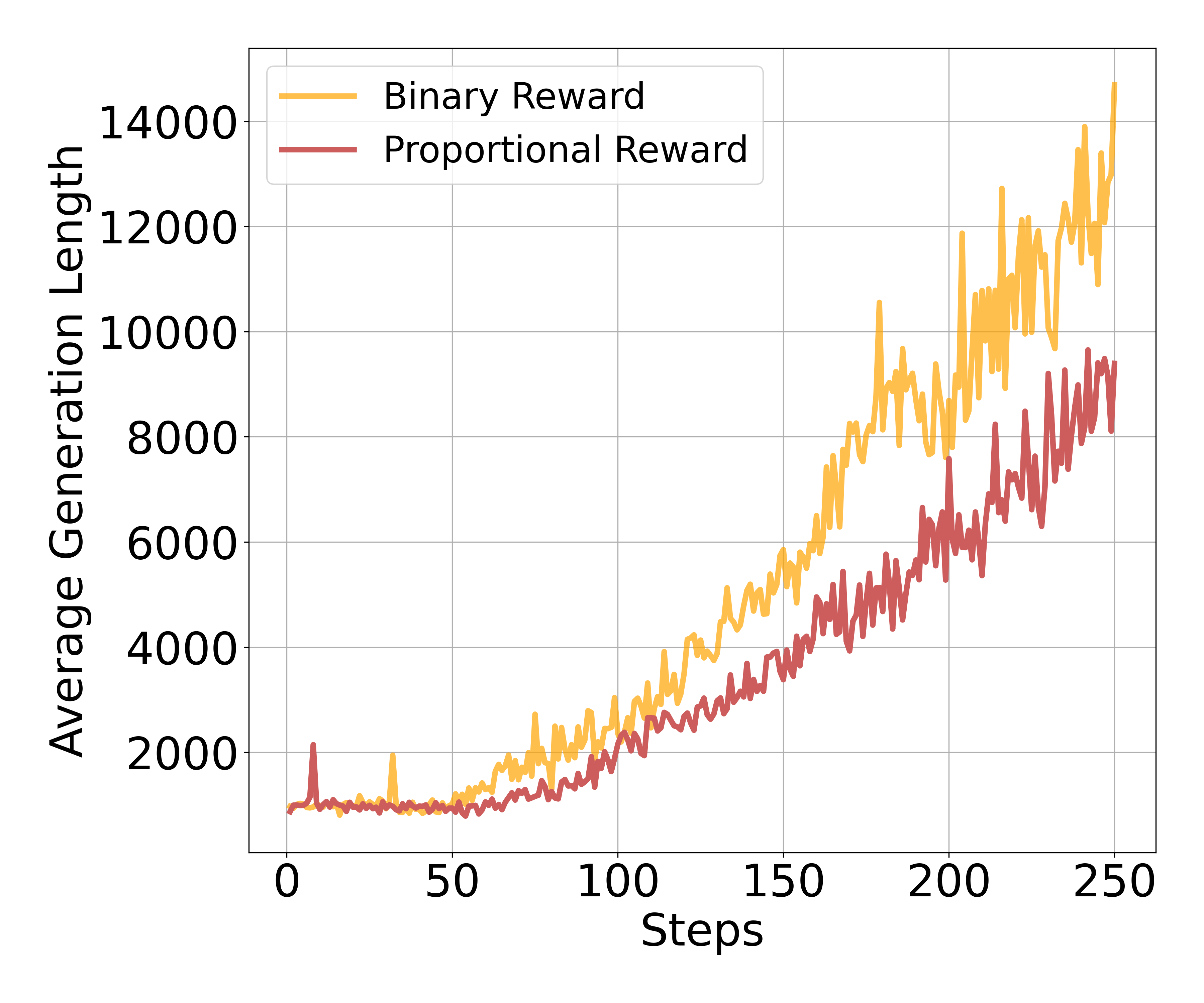}
    \end{minipage}
    \caption{
        \label{fig:partial_reward_ablation}
        \looseness=-1
        \textbf{Binary vs proportional reward for code problems.}
        (a) Accuracy on AIME and LiveCodeBench after 250 steps of training with binary reward and proportional reward. Performance on LiveCodeBench is 2\% lower with proportional rewards.
        (b) Length evolution throughout training. Length increases more with binary rewards.
    }
\end{figure}

To address this, we experimented with a proportional reward: based on the fraction of tests passed, as opposed to the binary reward discussed in Section~\ref{subsec:correctness}. In an ablation with a 24B model over 250 steps, we found that training with proportional rewards was faster, discarding three times less data. However, this approach led to slightly lower final performance on benchmarks, with a 2\% decrease on LiveCodeBench (Figure~\ref{fig:partial_reward_ablation}a), and slower growth in generation length (Figure~\ref{fig:partial_reward_ablation}b).

The hope was that a reward based on the fraction of tests passed should provide a richer signal than a simple pass/fail for RL training. However, the potential issue is that partial rewards could also provide false signal to incorrect solutions and be more sensitive to minor inconsistencies between implementations, potentially leading to less meaningful training batches.

\subsubsection{Entropy targeting}

\begin{figure}[htbp]
    \centering
    \begin{minipage}{0.48\textwidth}
        \centering
        \includegraphics[width=\linewidth]{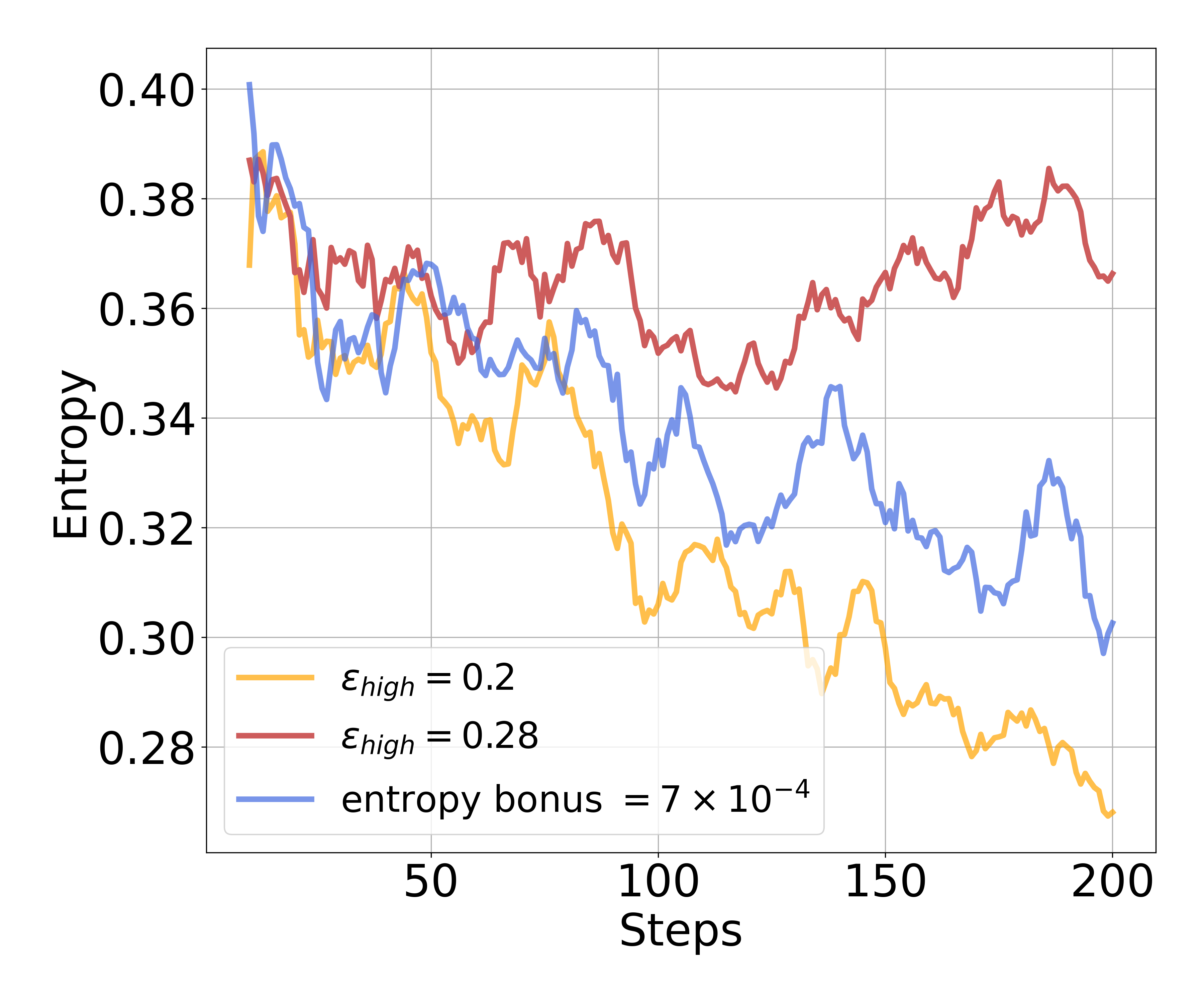}
    \end{minipage}\hfill
    \begin{minipage}{0.48\textwidth}
        \centering
        \includegraphics[width=\linewidth]{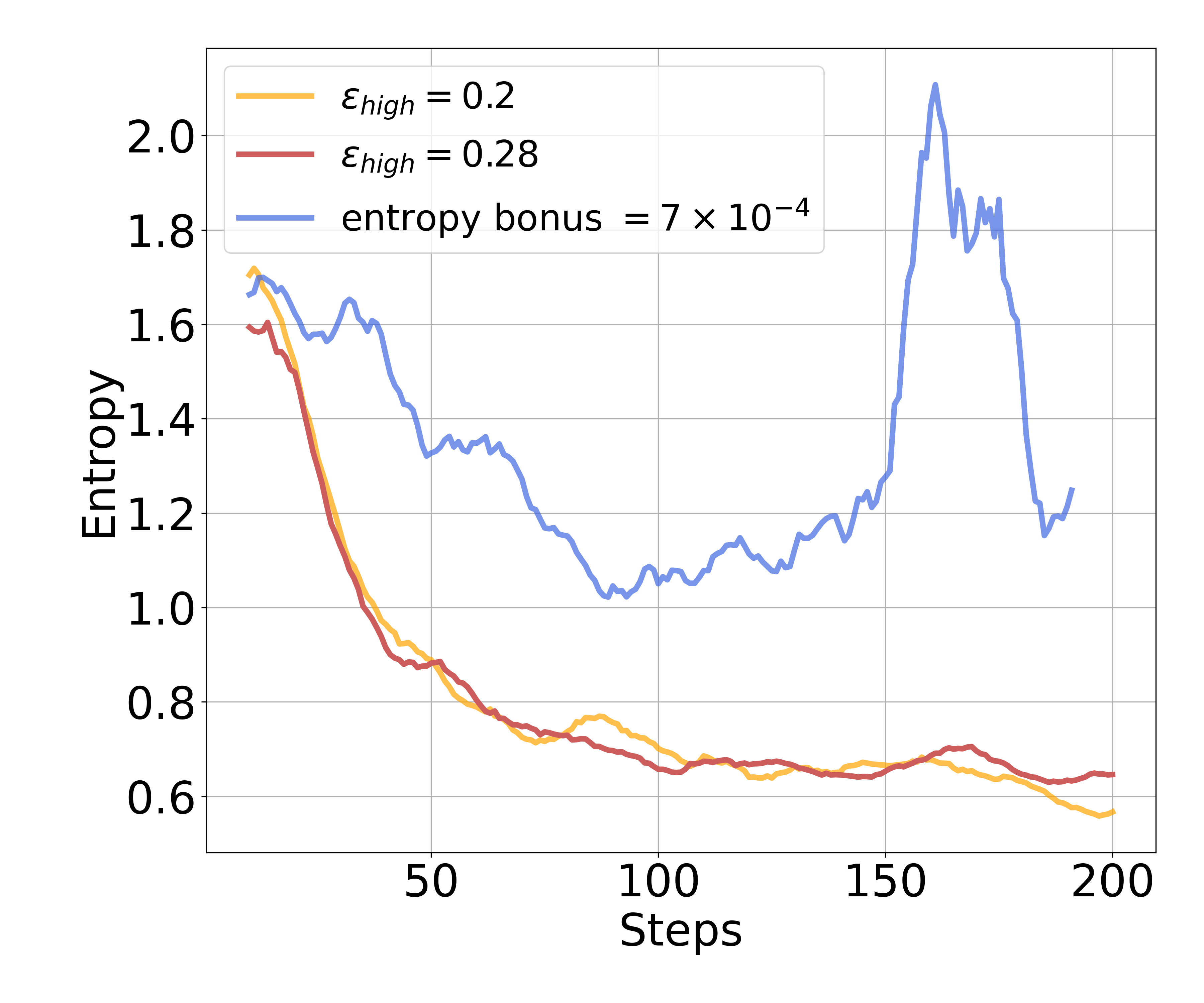}
    \end{minipage}
    \caption{
        \label{fig:entropy_ablation}
        \looseness=-1
        \textbf{Impact of $\varepsilon_{\text{high}}$ on the entropy distribution throughout training.}
        (a) Entropy evolution throughout training of a 3B model on a \textbf{math only} dataset. Entropy drops with entropy bonus, while higher $\varepsilon_{\text{high}}$ maintains entropy, allowing for better exploration. 
        (b) Entropy evolution throughout training of a 3B model on a \textbf{math and code} dataset. Entropy explodes with entropy bonus, even though the coefficient is the same as the math only version. Higher $\varepsilon_{\text{high}}$ behaves better, allowing entropy to decrease.
    }
\end{figure}

To encourage exploration and prevent entropy collapse during RL training, a common strategy in the literature~\citep{schulman2017proximal} is to add an entropy bonus loss term. However, we found this strategy to be unstable as the effect of the entropy bonus varies significantly depending on the dataset. For a math-only dataset, entropy drops with the entropy bonus, while a higher $\varepsilon_{\text{high}}$ maintains entropy, enhancing exploration (Figure~\ref{fig:entropy_ablation}a). On a math and code dataset, entropy increases excessively with the entropy bonus (even with the same coefficient as in the math-only run), while a higher $\varepsilon_{\text{high}}$ allows entropy to decrease, improving exploitation (Figure~\ref{fig:entropy_ablation}b).

Instead, we found it more effective to depend on $\varepsilon_{\text{high}}$, as also noted in literature~\citep{yu2025dapo, wang2025beyond}. This method avoids the instability issues associated with entropy bonuses.

Another approach for controlling entropy is adding a KL term to the PPO loss. However, as the generation distribution is expected to deviate significantly from the original model, we found that using a KL penalty primarily hinders training, consistent with previous findings~\citep{yu2025dapo}. We attempted using an exponential moving average of the weights during training as a reference for KL, but found it simpler to manually adjust $\varepsilon_{\text{high}}$.

\begin{figure}[H]
    \centering
    \includegraphics[width=0.9\textwidth]{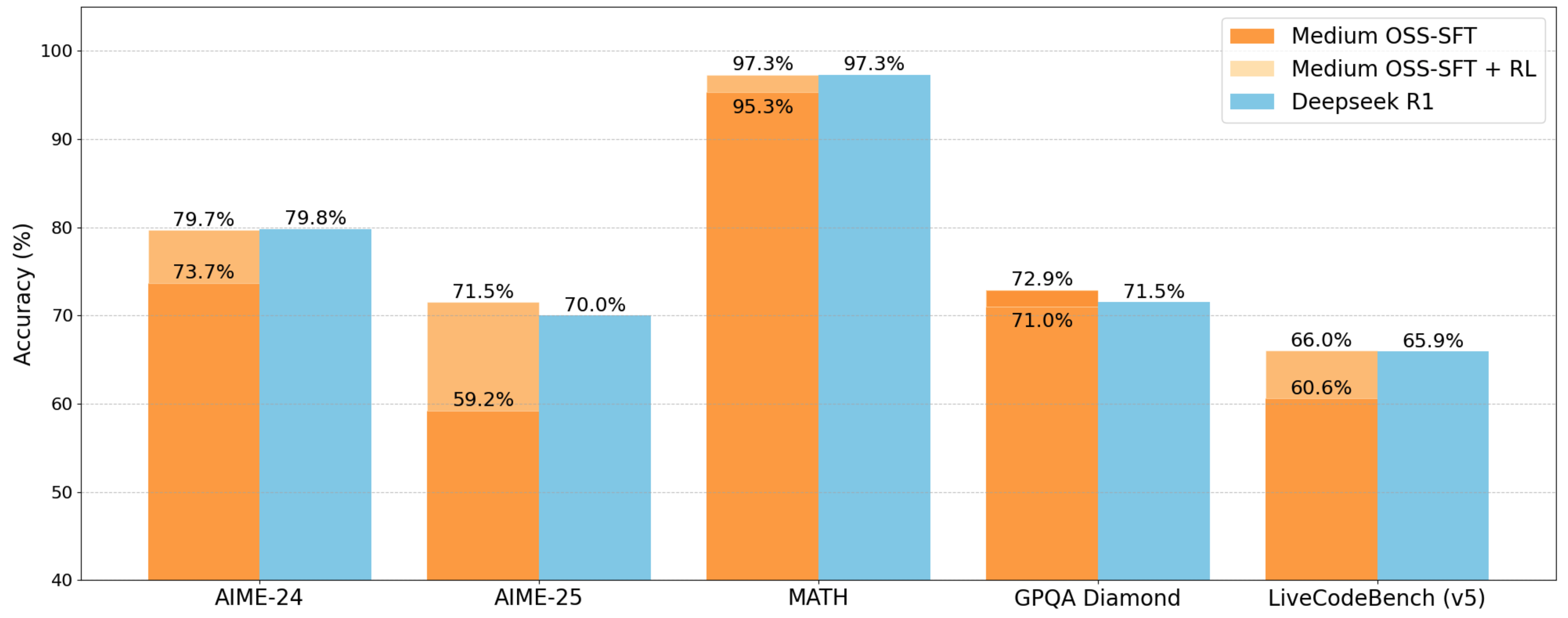}
    \caption{
    \textbf{Benchmark performance of Magistral Medium fine-tuned on open-source traces.}
    All results are reported using pass@1.
    The shaded region highlights the additional improvement achieved through RL on top of supervised fine-tuning.
    We find that while fine-tuning on open-source traces yields strong results, applying RL further enhances performance significantly. In particular, the accuracy on AIME'25 increases by more than 12\%. Please note that the performance on GPQA Diamond drops after RL from 72.9\% to 71.0\%.
    \label{fig:magistral-oss-rl}
    }
\end{figure}
\section{RL on model finetuned using OSS reasoning traces}
\label{sec:rl_oss}

As an experiment, we also tried to first finetune Mistral~Medium~3 using open source reasoning datasets OpenThoughts~\citep{guha2025openthoughtsdatarecipesreasoning} and the code subset of OpenR1~\citep{openr1, penedo2025codeforces} including both the prompts and the generations from these datasets i.e. Deepseek R1 generated traces. This included a total of about 1.3M generations. We then run RL on top of this finetuned checkpoint using our most difficult subset of the data.
As shown in Figure~\ref{fig:magistral-oss-rl}, applying RL yields substantial performance gains over the SFT checkpoint. Notably, the RL model improves by over 10 points on AIME’25 and 5 points on LiveCodeBench, achieving a final performance level on par with Deepseek-R1 on code and math benchmarks.

\section{Conclusion}
\label{sec:conclusion}

Magistral is our first step towards generally capable systems with reinforcement learning.
We look forward to the next research problems ahead of us: what loss and optimization algorithms are the most appropriate, how much gain can be unlocked by bootstrapping a model on its own reasoning traces, or how to scale to the next order of magnitude of compute. Looking ahead, we are also excited to push the boundaries of RL across a whole range of applications, with tool-use, integrated multimodality, and agents. As we explore this frontier, we remain committed to contributing to science in a transparent and optimistic manner.

\clearpage

\subsection*{Core contributors}

Abhinav Rastogi, Albert Q. Jiang, Andy Lo, Gabrielle Berrada, Guillaume Lample, Jason Rute, Joep Barmentlo, Karmesh Yadav, Kartik Khandelwal, Khyathi Raghavi Chandu, Léonard Blier, Lucile Saulnier, Matthieu Dinot, Maxime Darrin, Neha Gupta, Roman Soletskyi, Sagar Vaze, Teven Le Scao, Yihan Wang

\subsection*{Contributors}

Adam Yang, Alexander H. Liu, Alexandre Sablayrolles, Amélie Héliou, Amélie Martin, Andy Ehrenberg, Anmol Agarwal, Antoine Roux, Arthur Darcet, Arthur Mensch, Baptiste Bout, Baptiste Rozière, Baudouin De Monicault, Chris Bamford, Christian Wallenwein, Christophe Renaudin, Clémence Lanfranchi, Darius Dabert, Devon Mizelle, Diego de las Casas, Elliot Chane-Sane, Emilien Fugier, Emma Bou Hanna, Gauthier Delerce, Gauthier Guinet, Georgii Novikov, Guillaume Martin, Himanshu Jaju, Jan Ludziejewski, Jean-Hadrien Chabran, Jean-Malo Delignon, Joachim Studnia, Jonas Amar, Josselin Somerville Roberts, Julien Denize, Karan Saxena, Kush Jain, Lingxiao Zhao, Louis Martin, Luyu Gao, Lélio Renard Lavaud, Marie Pellat, Mathilde Guillaumin, Mathis Felardos, Maximilian Augustin, Mickaël Seznec, Nikhil Raghuraman, Olivier Duchenne, Patricia Wang, Patrick von Platen, Patryk Saffer, Paul Jacob, Paul Wambergue, Paula Kurylowicz, Pavankumar Reddy Muddireddy, Philomène Chagniot, Pierre Stock, Pravesh Agrawal, Romain Sauvestre, Rémi Delacourt, Sanchit Gandhi, Sandeep Subramanian, Shashwat Dalal, Siddharth Gandhi, Soham Ghosh, Srijan Mishra, Sumukh Aithal, Szymon Antoniak, Thibault Schueller, Thibaut Lavril, Thomas Robert, Thomas Wang, Timothée Lacroix, Valeriia Nemychnikova, Victor Paltz, Virgile Richard, Wen-Ding Li, William Marshall, Xuanyu Zhang, Yunhao Tang

\clearpage

\bibliography{ref}
\vfill
\pagebreak

\appendix

\begin{figure*}[t!]
\centering
\begin{tcolorbox}[
    colback=white,      
    colframe=black,     
    boxrule=1pt,        
    sharp corners       
]

    \noindent\textbf{Problem:}

    \includegraphics[width=0.4\linewidth]{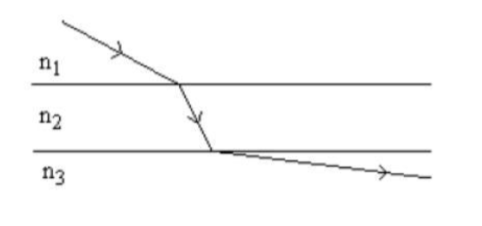}

A beam of light passes from medium 1 to medium 2 to medium 3 as shown in the diagram. What may be concluded about the speed of light in each medium?
    \par A. $v_3 > v_1 > v_2$
    \par B. $v_1 > v_2 > v_3$
    \par C. $v_1 > v_3 < v_2$ 
    \par D. $v_2 > v_3 > v_1$

    \vspace{0.5cm}
    \hrule
    \vspace{0.5cm}

    \noindent\textbf{Model Generation:}

    \vspace{0.2cm}

 {\ttfamily\scriptsize
<think> First, I need to understand the behavior of light as it passes through different mediums. According to Snell's Law, the speed of light changes when it enters a medium with a different refractive index. The refractive index ($n$) of a medium is inversely related to the speed of light ($v$) in that medium, meaning that as the refractive index increases, the speed of light decreases.
\par\medskip
Looking at the diagram:
The light beam starts from medium 1 to medium 2, bending towards the normal line. This indicates that medium 2 has a higher refractive index than medium 1 because light bends towards the normal when it slows down (entering a denser medium).
The light beam that passes from medium 2 to medium 3, bending away from the normal line. This indicates that medium 3 has a lower refractive index than medium 2 because light bends away from the normal when it speeds up (entering a less dense medium).
\par\medskip
Now, let's analyze the speed of light in each medium:
In medium 1, the speed of light is $v_1$.
In medium 2, since the light bends towards the normal when entering from medium 1, medium 2 has a higher refractive index than medium 1. Therefore, the speed of light in medium 2 is $v_2$, is less than in medium 1, so $v_1\,>\,v_2$.
In medium 3, since the light bends away from the normal when entering from medium 2, medium 3 has a lower refractive index than medium 2. Therefore, the speed of light in medium 3, $v_3$, is greater than in medium 2, so $v_3\,>\,v_2$.
\par\medskip
Now, we need to compare the speeds $v_1$, $v_2$, and $v_3$. From the above analysis,
$v_1\,>\,v_2$
$v_3\,>\,v_2$
But we also need to determine the relationship between $v_1$ and $v_3$. Since medium 3 has a lower refractive index than medium 1(because light bends away from the normal when transitioning from medium 2 to medium 3, and we already know medium 2 has a higher refractive index than medium 1), it implies that medium 3 is less dense than medium 1. Therefore, the speed of light in medium 3 will be greater than in medium 1.
\par\medskip
So, in sum: $v_3\,>\,v_1\,>\,v_2$
\par\medskip
Looking at the options provided: A. $v_3\,>\,v_1\,>\,v_2$ B. $v_1\,>\,v_2\,>\,v_3$ C. $v_1\,>\,v_3\,<\,v_2$ D. $v_2\,>\,v_3\,>\,v_1$
The correct option that matches our conclusion is A.
\par\medskip
</think>
}

    \vspace{0.5cm}
    
    \noindent Final Answer: A

\end{tcolorbox} 
\caption{A physics multimodal problem and its solution generated by Magistral Medium.}
\label{fig:mm-physics}
\end{figure*}

\begin{figure*}[t!]
\centering
\begin{tcolorbox}[
    colback=white,      
    colframe=black,     
    boxrule=1pt,        
    sharp corners       
]

    \noindent\textbf{Problem:}

    Which arrow points to a hydrogen bond?

    \includegraphics[width=0.4\linewidth]{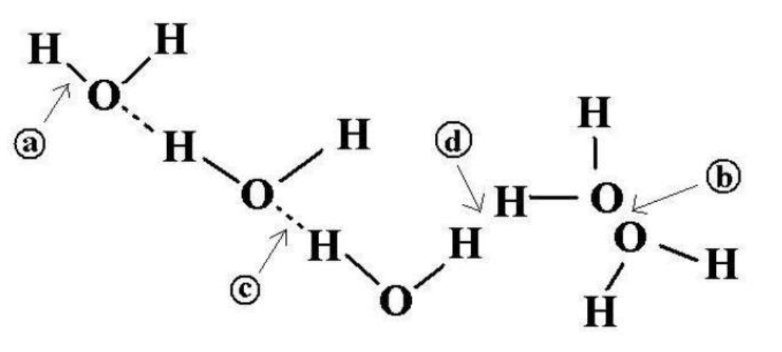}

    \vspace{0.5cm}
    \hrule
    \vspace{0.5cm}

    \noindent\textbf{Model Generation:}

    \vspace{0.2cm}

    {\ttfamily\scriptsize
    <think> Alright, I have this image showing a molecular structure with arrows labeled a, b, c, and d. The question is asking which arrow points to a hydrogen bond. Let's break this down.
    \par\medskip
    First, I need to understand what a hydrogen bond is. A hydrogen bond is a type of attractive interaction between a hydrogen atom and an electronegative atom like oxygen, nitrogen, or fluorine. In this case, since we have water molecules (H2O), the hydrogen bonds would be between the hydrogen atoms and the oxygen atoms of different water molecules.
    \par\medskip
    Looking at the image:
    \begin{itemize}[nosep, leftmargin=15pt]
        \item Arrow a points to a bond between a hydrogen atom and an oxygen atom within the same water molecule. That's a covalent bond, not a hydrogen bond.
        \item Arrow b points to another bond within a single water molecule, which is again a covalent bond.
        \item Arrow c points to a bond between a hydrogen atom of one water molecule and an oxygen atom of another water molecule. This looks like a hydrogen bond because it's between different water molecules.
        \item Arrow d points to a bond between hydrogen and oxygen within a single water molecule, so that's another covalent bond.
    \end{itemize}
    \par\medskip
    From this, it seems that arrow c is pointing to the hydrogen bond between the hydrogen of one water molecule and the oxygen of another water molecule.
    \par\medskip
    But just to be sure, I should double check. Hydrogen bonds typically occur between molecules, not within the same molecule, and they involve a hydrogen atom being attracted to a highly electronegative atom like oxygen. Arrow c is indeed showing this intermolecular interaction between an H of one molecule and O of another molecule.
    \par\medskip
    </think>
    }

    \vspace{0.5cm}

    \noindent The image shows four arrows labeled a, b, c, and d. Arrow c points to a bond between a hydrogen atom in one water molecule and an oxygen atom in another water molecule. This is characteristic of a hydrogen bond, which occurs between different molecules, not within the same molecule.
    
    \vspace{0.5cm}
    
    \noindent Final Answer: c

\end{tcolorbox} 
\caption{A chemistry multimodal problem and its solution generated by Magistral Medium.}
\label{fig:mm-chemistry}
\end{figure*}

\begin{figure*}[t!]
\centering
\begin{tcolorbox}[
    colback=white,      
    colframe=black,     
    boxrule=1pt,        
    sharp corners       
]

    \noindent\textbf{Problem:}

\noindent What is leading to the crinkling of this leaf's veins?
\par A. Bacterial pathogen 
\par B. I don't know and I don't want to guess 
\par C. Fungal pathogen 
\par D. Oomycete (watermould) pathogen 
\par E. Physiological condition; no pathogen involved

    \includegraphics[width=0.4\linewidth]{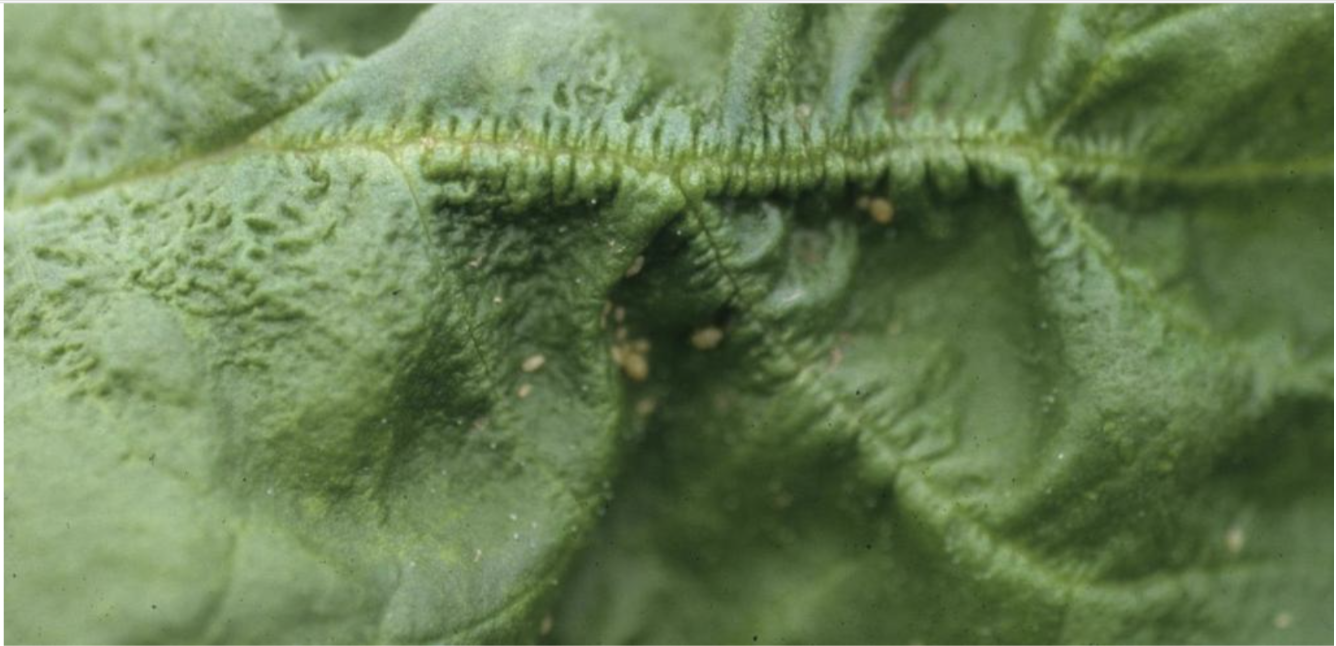}

    \vspace{0.5cm}
    \hrule
    \vspace{0.5cm}

    \noindent\textbf{Model Generation:}

    \vspace{0.2cm}

{\ttfamily\scriptsize
<think> The image shows a close-up of a leaf with some crinkling and discoloration along the veins. To determine the cause of the crinkling, we need to consider the different possibilities given in the options.
\par\medskip
Firstly, let's rule out some options:
Bacterial pathogens (Option A) usually cause symptoms such as water-soaked lesions, wilting, and bacterial ooze, not typically the crinkling seen here.
Fungal pathogens (Option C) often cause spots, mildews, or powdery substances on leaves, which doesn't match the appearance here.
Oomycete (watermould) pathogens (Option D) often lead to water-soaked lesions and a more generalized blight rather than vein-specific crinkling.
\par\medskip
Now, let's consider the remaining options:
Option B ("I don't know and I don't want to guess") is not a valid answer as we are tasked to determine the cause of the crinkling.
Option E (physiological condition) could be a possibility as it involves conditions like nutrient deficiencies or environmental stress that can lead to crinkling without pathogen involvement.
\par\medskip
The appearance of the leaf shows crinkling and some discoloration along the veins, which can often be attributed to physiological conditions. These might include nutrient deficiencies, water stress, or other environmental factors that do not involve pathogens.
\par\medskip
Given the visual clues and the typical symptoms associated with different causes, the most likely cause of the crinkling in this leaf's veins is a physiological condition rather than a pathogen.
\par\medskip
</think>
}

    \vspace{0.5cm}

    \noindent Summary: The image shows a leaf with crinkled veins and some discoloration. After considering the typical symptoms associated with various causes, it's most likely that the crinkling is due to a physiological condition rather than a bacterial, fungal, or oomycete pathogen. Therefore, the correct answer is:
    
    \vspace{0.5cm}
    
    \noindent Final Answer: E

\end{tcolorbox} 
\caption{A biology multimodal problem and its solution generated by Magistral Medium.}
\label{fig:mm-biology}
\end{figure*}

\end{document}